\documentclass[lettersize,journal]{IEEEtran}
\usepackage{amsfonts,amsmath,amssymb} % Math fonts and symbols
\usepackage{bbding}
\usepackage{nicefrac} % Compact fractions
\usepackage{graphicx} % Graphics
\usepackage[export]{adjustbox} % Extended graphic options
\usepackage{url} % URL formatting
\usepackage[nocompress]{cite} % Sorted, compressed citations
\usepackage{xcolor} % Color text and links
\usepackage{hyperref} % Hyperlinks
\hypersetup{
	colorlinks=true,
	linkcolor=blue,
	filecolor=red,      
	urlcolor=blue,
	citecolor=blue,
}
\usepackage{float} % Precise float control
\usepackage{wrapfig} % Text wrap around figures
\usepackage{stfloats}% Support for double-column floats

\usepackage{subcaption}
\usepackage{wasysym}
\usepackage{microtype} % 排版微调（字距、连字）

\usepackage{array}
\usepackage{booktabs}
\usepackage{colortbl}
\usepackage{makecell}
\usepackage{multirow}

\usepackage{algorithm}
\usepackage{algpseudocode}

\usepackage[capitalise]{cleveref}
\usepackage{orcidlink}

\usepackage{textcomp}
\usepackage{verbatim}
\usepackage{tikz}

\usepackage{pgfplots}
\pgfplotsset{compat = 1.18}
\usepackage{geometry}
\begin{document}

\title{HAD: Hierarchical Asymmetric Distillation to Bridge Spatio-Temporal Gaps in Event-Based Object Tracking}

\author{
% IEEE Publication Technology,~\IEEEmembership{Staff,~IEEE,}

Yao~Deng\orcidlink{0009-0006-0551-6294},
Xian~Zhong\orcidlink{0000-0002-5242-0467},~\IEEEmembership{Senior~Member,~IEEE},
Wenxuan~Liu\orcidlink{0000-0002-4417-6628},~\IEEEmembership{Member,~IEEE},
Zhaofei~Yu\orcidlink{0000-0002-6913-7553},~\IEEEmembership{Member,~IEEE},
Jingling~Yuan\orcidlink{0000-0001-7924-8620},~\IEEEmembership{Senior~Member,~IEEE},
and~Tiejun~Huang\orcidlink{0000-0002-4234-6099},~\IEEEmembership{Senior~Member,~IEEE}

% <-this % stops a space
 
% \thanks{This paper was produced by the IEEE Publication Technology Group. They are in Piscataway, NJ.}% <-this % stops a space
% \thanks{Manuscript received April 19, 2021; revised August 16, 2021.}}

\thanks{Manuscript Received October 18, 2025.
This work was supported by the National Natural Science Foundation of China (Grants No. 62472332 and 62271361), the Hubei Provincial Key Research and Development Program (Grant No. 2024BAB039), and the Hubei Key Laboratory of Inland Shipping Technology (Grant No. NHHY2024003). The authors acknowledge Beijing PARATERA Technology Co., LTD for providing high-performance and AI computing resources. (\textit{Corresponding authors: Xian Zhong.})
}

\thanks{Yao Deng, Xian Zhong, and Jingling Yuan are with the Sanya Science and Education Innovation Park, Wuhan University of Technology, Sanya 572025, China, and also with the Hubei Key Laboratory of Transportation Internet of Things, School of Computer Science and Artificial Intelligence, Wuhan University of Technology, Wuhan 430070, China (e-mail: 361248@whut.edu.cn; zhongx@whut.edu.cn; yjl@whut.edu.cn).}

\thanks{Wenxuan Liu, Zhaofei Yu, and Tiejun Huang are with the State Key Laboratory for Multimedia Information Processing, Peking University, Beijing 100091, China (e-mail: liuwx66@pku.edu.cn; yuzf12@pku.edu.cn; tjhuang@pku.edu.cn).}

}

% \author{
% % IEEE Publication Technology,~\IEEEmembership{Staff,~IEEE,}

% Yao~Deng\orcidlink{0009-0006-0551-6294},
% Xian~Zhong\orcidlink{0000-0002-5242-0467},~\IEEEmembership{Senior~Member,~IEEE},
% Wenxuan~Liu\orcidlink{0000-0002-4417-6628},~\IEEEmembership{Member,~IEEE},
% Zhaofei~Yu\orcidlink{0000-0002-6913-7553},~\IEEEmembership{Member,~IEEE},
% Jingling~Yuan\orcidlink{0000-0001-7924-8620},~\IEEEmembership{Senior~Member,~IEEE},
% and~Tiejun~Huang\orcidlink{0000-0002-4234-6099},~\IEEEmembership{Senior~Member,~IEEE}

% \thanks{Yao Deng and Jingling Yuan are with the Sanya Science and Education Innovation Park, Wuhan University of Technology, China, and also with the Hubei Key Laboratory of Transportation Internet of Things, School of Computer Science and Artificial Intelligence, Wuhan University of Technology, China (e-mail: 361248@whut.edu.cn; yjl@whut.edu.cn).}

% \thanks{Xian Zhong is with the Hubei Key Laboratory of Transportation Internet of Things, School of Computer Science and Artificial Intelligence, Wuhan University of Technology, China, and also with the State Key Laboratory of Maritime Technology and Safety, Wuhan University of Technology, China (e-mail: zhongx@whut.edu.cn).}

% \thanks{Wenxuan Liu, Zhaofei Yu, and Tiejun Huang are with the State Key Laboratory for Multimedia Information Processing, Peking University, China (e-mail: liuwx66@pku.edu.cn; yuzf12@pku.edu.cn; tjhuang@pku.edu.cn).}

% }

\maketitle

\begin{abstract}
RGB cameras excel at capturing rich texture details with high spatial resolution, whereas event cameras offer exceptional temporal resolution and a high dynamic range (HDR). Leveraging their complementary strengths can substantially enhance object tracking under challenging conditions, such as high-speed motion, HDR environments, and dynamic background interference. However, a significant spatio-temporal asymmetry exists between these two modalities due to their fundamentally different imaging mechanisms, hindering effective multi-modal integration. To address this issue, we propose {Hierarchical Asymmetric Distillation} (HAD), a multi-modal knowledge distillation framework that explicitly models and mitigates spatio-temporal asymmetries. Specifically, HAD proposes a hierarchical alignment strategy that minimizes information loss while maintaining the student network’s computational efficiency and parameter compactness. Extensive experiments demonstrate that HAD consistently outperforms state-of-the-art methods, and comprehensive ablation studies further validate the effectiveness and necessity of each designed component. The code will be released soon.
% at \url{https://github.com/Taxalfer/HAD}.

\end{abstract}


\section{Introduction}

%事件跟踪的背景 现有的问题是什么

\begin{figure}[!t]
	\centering
	\includegraphics[width = \linewidth]{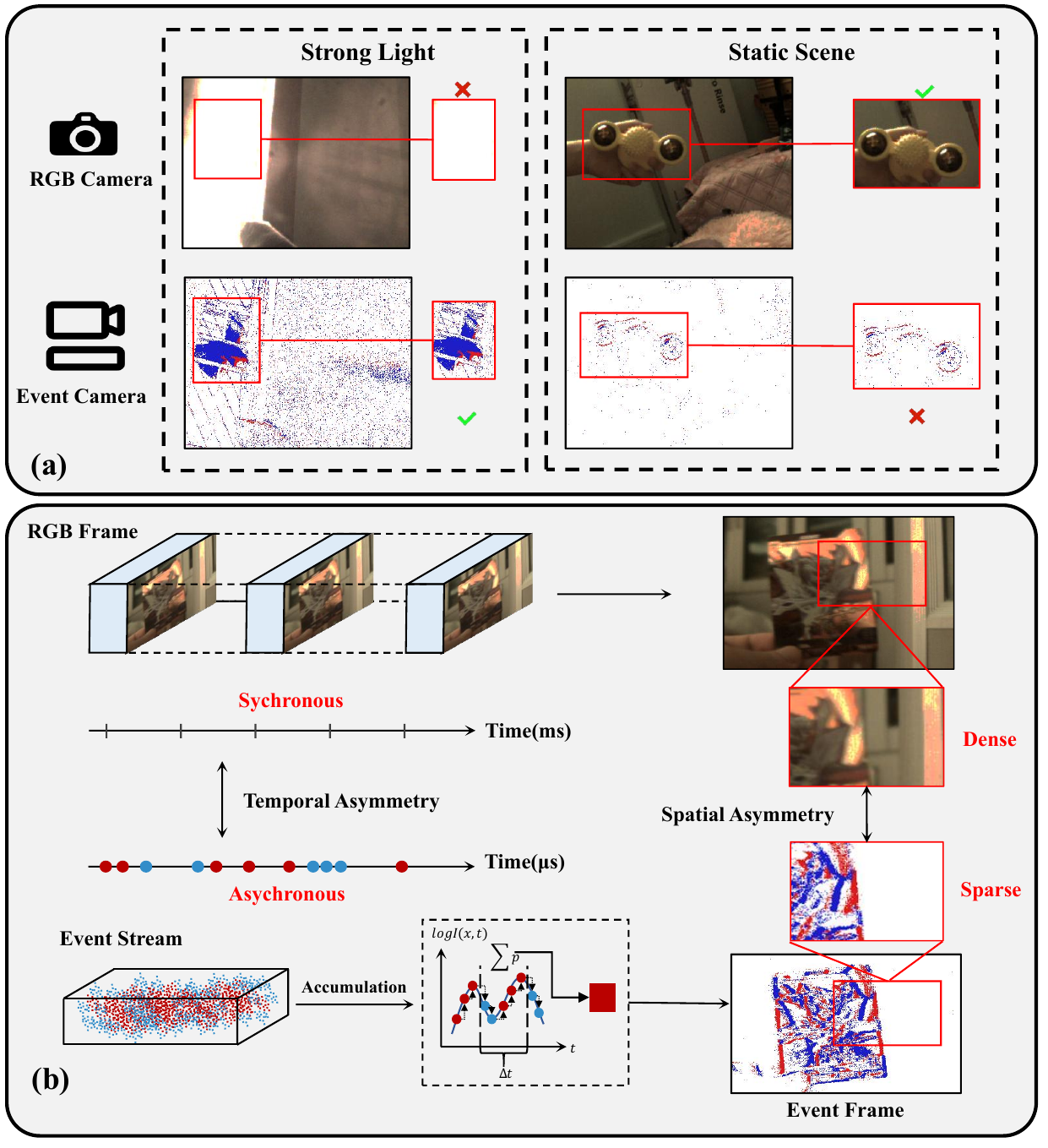}
	\caption{\textbf{Motivation of HAD.} 
	(a) RGB cameras capture rich texture details under standard conditions, whereas event cameras encode rapid motion information under extreme environments. Multi-modal fusion leverages these complementary advantages. 
	(b) Effective fusion requires explicitly addressing the inherent temporal and spatial asymmetry between RGB frames and event streams.}
	\label{fig:motivation}
\end{figure}

\IEEEPARstart{E}{vent} cameras represent a revolutionary advancement in visual sensing technology. Unlike traditional frame-based cameras, event cameras operate in an event-driven manner: each pixel asynchronously detects luminance changes and generates discrete events with precise timestamps and polarity information, indicating increases or decreases in brightness~\cite{pami/GallegoDOBTCLDC22, jssc/LichtsteinerPD08}. This sensing paradigm offers several distinct advantages. First, event cameras capture dynamic scenes with exceptional temporal resolution, often on the order of microseconds, enabling ultra-low-latency perception of rapid motion, which is highly suitable for real-time applications such as robotics and autonomous driving. Second, they exhibit a high dynamic range (HDR), allowing robust recording under extreme brightness variations without over- or underexposure. Third, their sparse data representation reduces computational complexity and storage requirements: since events are generated only when luminance changes occur, the output remains significantly sparser than continuous frame streams. Such sparsity is particularly advantageous for resource-constrained systems such as embedded devices and mobile robots.

Despite these merits, event cameras face notable challenges in object tracking. A primary limitation lies in the inherently sparse nature of event data: in static or slowly changing scenes, event density becomes extremely low, yielding insufficient information for robust tracking. This sparsity makes it difficult to maintain accurate and consistent target localization, restricting the modality’s applicability in real-world single-object tracking (SOT). Recent event-based tracking methods alleviate this issue by integrating event streams with complementary modalities such as RGB images or depth maps, thereby improving robustness and accuracy~\cite{zhang2023frame, wang2024event, wang2021viseventbenchmark}. However, most of these approaches assume modality consistency during distillation and overlook the intrinsic \textit{spatio-temporal asymmetry} between RGB frames and event streams. This oversight can cause severe misalignment, leading to suboptimal knowledge transfer and degraded tracking performance under challenging conditions (see \cref{fig:motivation}).

In SOT, accurate localization requires both \textit{precise appearance modeling} and \textit{robust temporal correspondence}. RGB frames provide abundant texture details but are temporally sparse, whereas event streams offer dense temporal cues but limited spatial texture. This spatio-temporal asymmetry can misalign semantic features when distilling from an RGB-event teacher to an event-only student. If the teacher emphasizes appearance cues absent in event data, the student may receive non-transferable supervision, leading to overfitting and degraded performance. This challenge raises two key issues.

\textit{Issue 1: Temporal Asymmetry.} Reconciling asynchronous temporal dynamics between the teacher and student is crucial. We propse a \textit{Temporal Alignment} (TA) module based on a lightweight Gated Recurrent Unit (GRU)~\cite{corr/ChungGCB14}. Both teacher and student feature sequences are processed through the GRU, which recursively updates hidden states to capture temporal dependencies. By aligning sequential information, the student benefits from richer historical context, enabling robust temporal modeling without dense frame-level supervision.

\textit{Issue 2: Spatial Asymmetry.} Another obstacle is the structural mismatch between RGB and event-based feature maps. Unlike RGB inputs, which retain fine-grained textures, event data primarily encode coarse structural cues. Accurate localization depends on maintaining \textit{consistent spatial response structures} rather than fine appearance details. To mitigate spatial distortion, we treat the teacher’s and student’s response maps as probability distributions and employ entropic-regularized optimal transport (OT)~\cite{monge1781histoire, kantorovich2006translocation} to compute a soft matching plan via Sinkhorn iterations~\cite{sinkhorn1967concerning}. This structure-aware OT loss respects the perceptual limitations of event data and avoids rigid one-to-one constraints.

Although multi-modal fusion methods can leverage complementary cues from RGB and event streams, they typically require both modalities during inference, which is impractical in real-world scenarios where RGB sensors may fail under extreme lighting (\textit{e.g.}, overexposure or low illumination). In contrast, knowledge distillation enables a unimodal (event-only) student to inherit robustness from a bimodal teacher during training, while maintaining low computational cost and modality independence at inference. This paradigm is particularly suitable for bridging spatio-temporal asymmetry: rather than enforcing direct feature fusion, distillation allows us to design alignment mechanisms (\textit{e.g.}, TA and SAOT) that selectively transfer the \textit{transferable knowledge}, temporal dynamics and structural spatial responses, while discarding non-transferable appearance details that event data cannot represent.

Building on these insights, we propose {Hierarchical Asymmetric Distillation} (HAD), a novel distillation framework explicitly designed to resolve the dual challenges of spatio-temporal asymmetry in event-based object tracking. Unlike generic multi-modal fusion methods, HAD is problem-driven: each component is directly motivated by, and tailored to, the specific facets of asymmetry we formally characterize.

Our main contributions are summarized threefold:

\begin{itemize}

 \item We formulate the intrinsic spatio-temporal asymmetry between RGB frames and event streams as two interdependent issues: temporal misalignment caused by asynchronous sampling and spatial mismatch arising from coarse event-driven representations.

 \item We design \textit{HAD}, a hierarchical distillation pipeline that directly addresses these two facets of asymmetry: (i) a Temporal Alignment (TA) module synchronizes temporal dynamics, and (ii) a Spatial-Aligned Optimal Transport (SAOT) module aligns response distributions while preserving structural consistency.

 \item We conduct extensive experiments on \textsc{EventVOT}, \textsc{COESOT}, and \textsc{VisEvent}, demonstrating that HAD achieves competitive performance against state-of-the-art fusion and distillation baselines, with strong robustness under noise, motion blur, and sparse inputs.

\end{itemize}

\section{Related Work}

\subsection{Neuromorphic Vision Sensors}

Neuromorphic vision sensors advance visual perception through bio-inspired mechanisms. Event cameras, also known as dynamic vision sensors (DVS)~\cite{pami/GallegoDOBTCLDC22}, operate asynchronously, generating pixel-level event streams with location, timestamp, and polarity information only when brightness changes occur. They achieve microsecond-level latency, a high dynamic range (HDR) exceeding 120~dB, and low power consumption. Since the Mead group introduced the silicon retina in the 1990s, commercial event cameras have achieved resolutions up to $1280 \times 720$~\cite{jssc/BrandliBYLD14} and have been applied to pose estimation~\cite{ijrr/MuegglerRGDS17}, motion segmentation~\cite{iros/MitrokhinYFAD19}, and object tracking~\cite{wang2024event, tang2022coesot}. 

The spike camera~\cite{corr/abs-2201-09302}, another asynchronous sensor, achieves ultra-high-speed imaging (1,000~FPS) via photon integration and spike modulation, emphasizing light-intensity accumulation and optical-flow estimation. The Asynchronous Event-Based Multikernel Algorithm~\cite{tnn/LagorceMIFB15} leverages event-driven sensors that capture only scene changes. By processing spatio-temporal events through an asynchronous framework, it enables high-precision tracking with low computational complexity, making it ideal for real-time, energy-efficient applications such as robot navigation, SLAM, and object recognition. Recent advances in event and spike cameras are expected to further drive progress in multi-modal perception~\cite{ijon/LiLGDZ25, cvpr/YanZCFLDSWXM024}.

\subsection{Multi-Modal Knowledge Distillation}

Multi-modal knowledge distillation has made remarkable progress in recent years. Scale-Decoupled Distillation (SDD)~\cite{cvpr/WeiLL24} introduced a scale-decoupling strategy to separate global logit outputs, improving distillation quality. In multi-modal cross-language video summarization (MCLS), a video-guided dual-fusion network (VDF) with a three-stage training strategy was developed to enhance summarization~\cite{pami/LiuWYTSYYJLF24}. U2MKD~\cite{pami/SunZTPQX24} addressed LiDAR-camera heterogeneity via bidirectional feature fusion and cross-modal transfer, while HDETrack~\cite{wang2024event} employed hierarchical distillation for efficient event-camera tracking. These studies highlight the strong potential of distillation for transferring complementary information across modalities. 

Furthermore, SinKD~\cite{tnn/CuiQGZXWLSZL25} employs the Sinkhorn distance with batching to more accurately measure and reduce distribution gaps between teacher and student models, mitigating mode-collapse issues. In contrast, our work focuses specifically on the spatio-temporal asymmetry between event streams and RGB frames, and proposes a dedicated distillation framework to enhance multi-modal representation learning for event-based tracking.

\subsection{Multi-Modal-Based Object Tracking}

Event cameras have greatly advanced object tracking by leveraging high temporal resolution, HDR, and low power consumption. TrDiMP~\cite{cvpr/0020ZWL21} was the first to propose the Transformer into visual tracking by decoupling the encoder and decoder into two parallel branches within a Siamese-like framework, using the encoder to enhance template features and the decoder to propagate temporal context, thereby improving tracking robustness. CrossEI~\cite{tip/ChenWDLS25} effectively aligns event and image modalities through a motion-adaptive event sampling strategy and a bidirectionally enhanced fusion framework, alleviating motion blur and background interference by incorporating image-guided motion estimation and semantic modulation. CSAM~\cite{nips/ZhangD0DH24} integrates multi-object tracking association with event-stream motion information, leveraging multi-modal fusion and spatio-temporal modeling in complex scenarios. 

MAFNet~\cite{tnn/LiuZLLT25} addresses appearance discrepancies caused by modality switching in cross-modal tracking by adaptively fusing features from RGB and NIR modalities. OSTrack~\cite{eccv/YeCMSC22} unifies template and search regions into a single one-stream framework with bidirectional feature flows, enabling end-to-end relation modeling and highly parallelized inference. AiATrack~\cite{eccv/GaoZMWY22} proposes an ``Attention-in-Attention'' (AiA) module that enhances Transformer discriminability and robustness through mutual negotiation among attention weights, achieving high-performance real-time tracking. SFTrack~\cite{corr/abs-2505-12903} employs a slow-fast dual-mode architecture for event streams, combining a high-precision slow path with a low-latency fast path to balance accuracy and efficiency in diverse deployment settings. 

Unlike prior methods, our HAD explicitly aligns the teacher’s dual-modality feature distributions, enabling the student network to exploit complementary cues more effectively and robustly.

\begin{figure*}[!t]
	\centering
	\includegraphics[width = 0.95\linewidth]{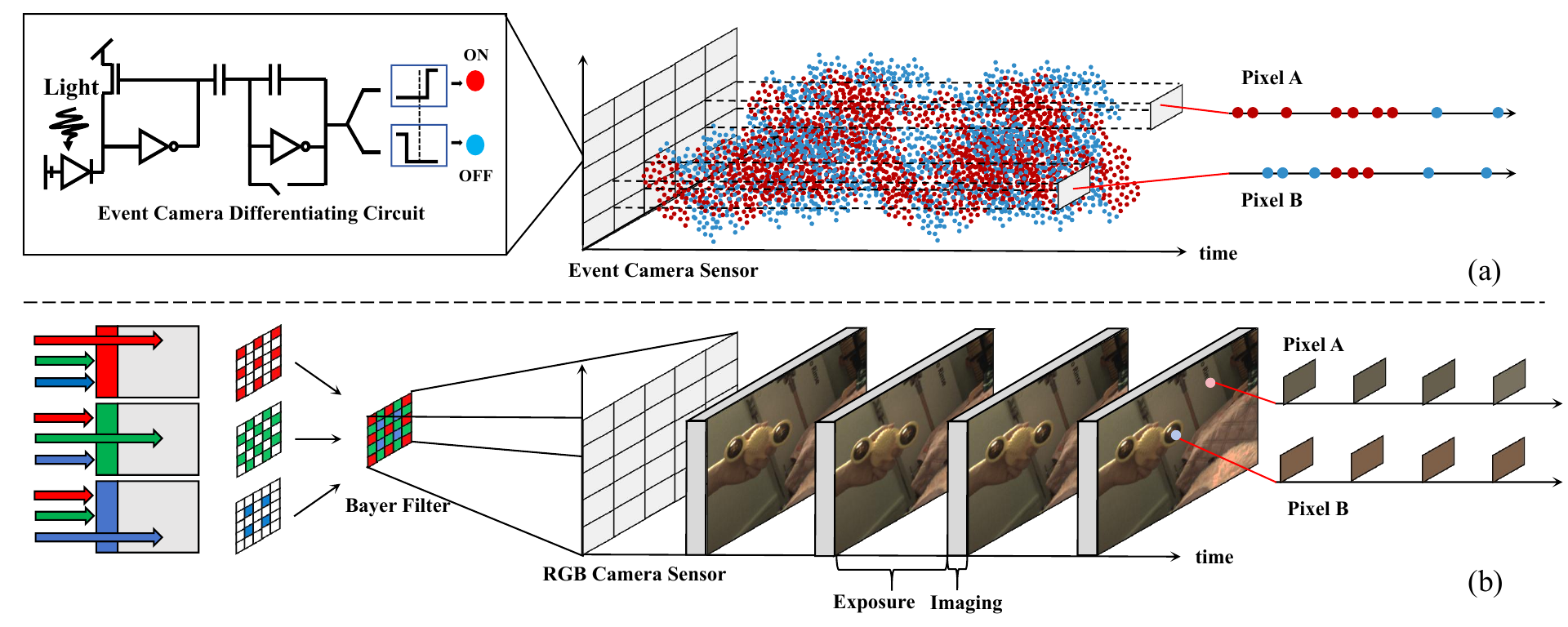}
	\caption{\textbf{Sampling mechanisms of RGB and event cameras.} 
	(a) Event cameras operate based on a differential circuit principle, asynchronously triggering ON/OFF events at each pixel in response to local light-intensity changes. 
	(b) RGB cameras employ a Bayer filter arrangement and perform synchronous sampling at a fixed frame rate, as used in traditional image sensors.}
	\label{fig:camera_mechanism}
\end{figure*}

\subsection{Optimal Transport}

Optimal Transport (OT)~\cite{monge1781histoire} originated from Monge’s ``sand-moving problem'', which sought the minimal-cost plan to transform one probability distribution into another. Over time, OT has evolved through Kantorovich’s linear-programming reformulation~\cite{kantorovich2006translocation}, Brenier’s gradient-mapping theory~\cite{brenier1991polar}, and Cuturi’s entropy-regularized algorithms~\cite{nips/Cuturi13}, becoming a versatile tool across geometry, optimization, and probability. 

In visual computing and signal processing, OT has emerged as a powerful method for distribution alignment. Wasserstein GANs~\cite{ArjovskyCB17} improved generative training stability by replacing Jensen-Shannon divergence with the Wasserstein distance. In cross-domain tasks, DAOT~\cite{zhu2023daot, tip/ZhongQZYHW25} employed dual-domain joint transport to align feature and geometric distributions for crowd counting. In distillation, SOTA~\cite{ijcai/abs-2505-00394} first integrated spike-camera temporal characteristics with OT to mitigate saliency-detection bias caused by noise. Building upon these insights, we propose the \textit{Spatial-Aligned OT} (SAOT) module to align high-dimensional feature distributions and response maps while respecting the sparse nature of event data.

\section{Proposed Method}

To effectively bridge the spatio-temporal asymmetry between RGB frames and event streams, we propose Hierarchical Asymmetric Distillation (HAD), a knowledge distillation framework that explicitly aligns multi-modal representations across both temporal and spatial domains. 

We first analyze the intrinsic differences in the sampling mechanisms of RGB and event cameras, which lead to the core challenge of modality misalignment (see \S\ref{sec:camera_sampling_mechanism}). Building on this analysis, we formalize the dual facets of spatio-temporal asymmetry and motivate a two-stage alignment strategy (see \S\ref{sec:asymmetry_analysis}). Finally, we present the overall HAD framework, which integrates a Temporal Alignment (TA) module to synchronize asynchronous feature dynamics and a {Spatial-Aligned Optimal Transport (SAOT)} module to align response distributions in a geometry-aware manner. This hierarchical design enables the event-only student network to effectively inherit the robustness of the bimodal teacher (see \S\ref{sec:had_framework}).

\subsection{Camera Sampling Mechanism}
\label{sec:camera_sampling_mechanism}

\cref{fig:camera_mechanism} illustrates the fundamental difference in sampling principles between RGB and event cameras, which directly leads to significant spatio-temporal asymmetry.

\textit{1) Event cameras} operate asynchronously. A pixel triggers an event at timestamp $t_k$ whenever the absolute change in logarithmic light intensity exceeds a preset threshold $C$:
\begin{align}
	\Delta \ell \left(x,y,t_k \right) = \log \frac{L \left(x,y,t_k \right)}{L \left(x,y,t_\mathrm{last} \right)} \ge \pm C,
\end{align}
where $\Delta \ell(x,y,t_k)$ denotes the log-radiance change since the last event at that pixel, $t_\mathrm{last}$ is the previous event timestamp, and $C$ is the contrast threshold (positive for ON events, negative for OFF events). 
This change-based triggering mechanism produces an uneven spatio-temporal distribution of events that encode only dynamic information. It effectively removes motion blur, precisely marks brightness changes, and minimizes redundancy in static regions.

\textit{2) RGB cameras} operate synchronously. At each fixed time $t_n$, a global exposure is applied to all pixels, and the intensity value $I(x,y,t_n)$ at pixel $(x,y)$ is given by: 
\begin{align}
	I \left(x,y,t_n \right) = \int_{t_n - \tau}^{t_n} L \left(x,y,t \right) \mathrm{d}t,
\end{align}
where $L(x,y,t)$ is the instantaneous radiance and $\tau$ is the exposure duration. 
Each frame integrates irradiance over $\tau$, resulting in temporal averaging. In dynamic scenes, this leads to motion blur. Furthermore, all pixels are sampled regardless of change, producing high redundancy.

\subsection{Spatio-Temporal Asymmetry Analysis}
\label{sec:asymmetry_analysis}

\subsubsection{Temporal Dimension}

Temporal performance is characterized by temporal resolution $\Delta t$, the smallest measurable interval, and end-to-end latency $\tau$, the delay from photon arrival to data output. 
For an RGB camera with frame rate $f$:
\begin{align}
	\Delta t_\mathrm{RGB} = \frac{1}{f}, \quad 
	\tau_\mathrm{RGB} = \tau_\mathrm{exp} + \frac{1}{f},
\end{align}
where $\Delta t_\mathrm{RGB}$ is the inter-frame interval, $\tau_\mathrm{RGB}$ is the total latency composed of exposure time $\tau_\mathrm{exp}$ and frame readout period $1/f$, and $f$ is the frame rate. 
By contrast, event cameras respond to per-pixel brightness changes with microsecond precision:
\begin{align}
	\Delta t_\mathrm{event} \sim \mathcal{O} \left(1 \mu s \right), \quad 
	\tau_\mathrm{event} \ll \tau_\mathrm{RGB},
\end{align}
where $\Delta t_\mathrm{event}$ is on the order of microseconds and $\tau_\mathrm{event}$ is typically less than 1 ms, enabling sub-millisecond latency and intrinsic immunity to motion blur, critical for high-speed perception.

\subsubsection{Spatial Dimension}

Spatial performance is described by spatial density $D$, the number of independent measurements per frame, and redundancy $R$, the degree of overlap in captured information. 
For an $N \times M$ RGB frame:
\begin{align}
	D_\mathrm{RGB} = N \times M, \quad 
	R_\mathrm{RGB} \approx 1,
\end{align}
where $D_\mathrm{RGB}$ represents the total number of sampled pixels, and $R_\mathrm{RGB} \approx 1$ indicates dense, redundant sampling, as most pixels in consecutive frames capture static backgrounds when scene dynamics are limited.

Event cameras employ sparse, data-driven sampling:
\begin{align}
	D_\mathrm{event} = K \left(t \right), \quad 
	R_\mathrm{event} \ll 1,
\end{align}
where $K(t)$ is the number of active pixels at time $t$, typically a small fraction of the total. Hence, $R_\mathrm{event} \ll 1$, meaning that events provide highly localized, low-redundancy information concentrated in dynamic regions.

Overall, RGB and event cameras exhibit complementary properties: RGB provides high spatial resolution but low temporal fidelity, whereas event cameras offer the opposite. This motivates explicit strategies to exploit and reconcile their asymmetry.

\begin{figure*}[!t]
	\centering
	\includegraphics[width = 0.95\linewidth]{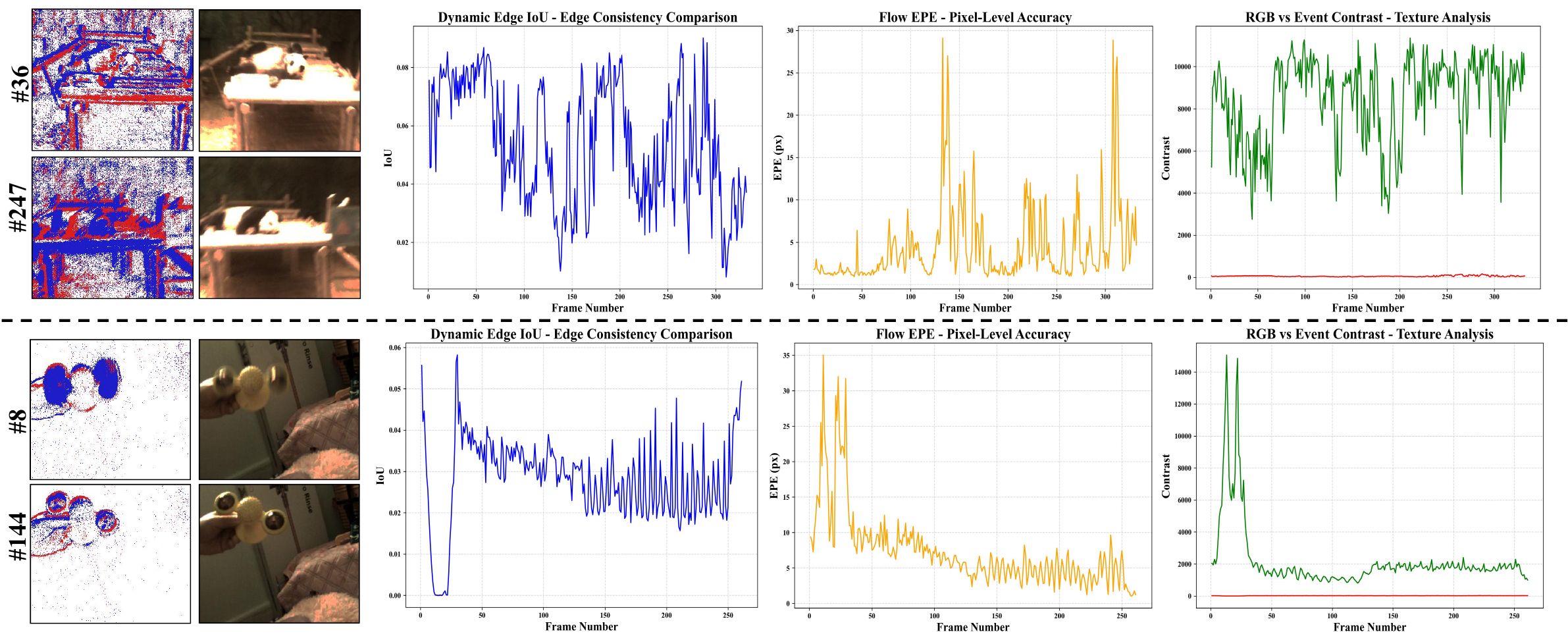}
	\caption{\textbf{Analysis of representative sequences on \textsc{COESOT}.} 
	Two representative sequences are separated by a dashed line. Each analysis panel (arranged from left to right) contains four complementary subfigures: 
	\textbf{(1)} Comparative visualization of RGB ground-truth keyframes and corresponding event frames; 
	\textbf{(2)} Dynamic-edge intersection-over-union (IoU) between RGB frames and event streams (\textcolor[HTML]{0000FF}{blue}); 
	\textbf{(3)} Optical-flow alignment errors for both modalities (\textcolor[HTML]{FFAC15}{yellow}); and 
	\textbf{(4)} Texture-contrast comparison between event streams (\textcolor[HTML]{007F00}{green}) and RGB frames (\textcolor[HTML]{F61D1D}{red}).}
	\label{fig:analysis}
\end{figure*}

\subsubsection{Discussion}

We analyze modality differences between RGB images and event frames across four metrics: dynamic edge detection~\cite{pami/Canny86a}, Intersection-over-Union (IoU)~\cite{ijcv/UijlingsSGS13}, texture contrast~\cite{tsmc/HaralickSD73}, and optical-flow endpoint error (EPE)~\cite{ai/HornS81}. In \cref{fig:analysis}, the left sequence depicts a simple scene with a rapidly moving object and a stationary camera, while the right sequence involves a complex background, slow-moving objects, and a rapidly shaking camera. 

During fast motion, RGB frames often suffer from severe blur (\textit{e.g.}, frame \#36 on the left and \#144 on the right) due to photon accumulation over the exposure time $\tau$. This blur substantially reduces edge IoU and increases optical-flow errors, degrading both edge preservation and motion estimation. In contrast, event frames maintain near-zero edge IoU while exhibiting high texture contrast, highlighting their sensitivity to dynamic structures. These observations reveal pronounced spatio-temporal disparities: RGB relies on inter-frame differences and is vulnerable to blur, whereas event frames capture brightness-change rates and respond rapidly to motion. Explicitly addressing these disparities is essential for robust cross-modal alignment in tracking tasks.

\begin{figure*}[!t]
	\centering
	\includegraphics[width = \linewidth]{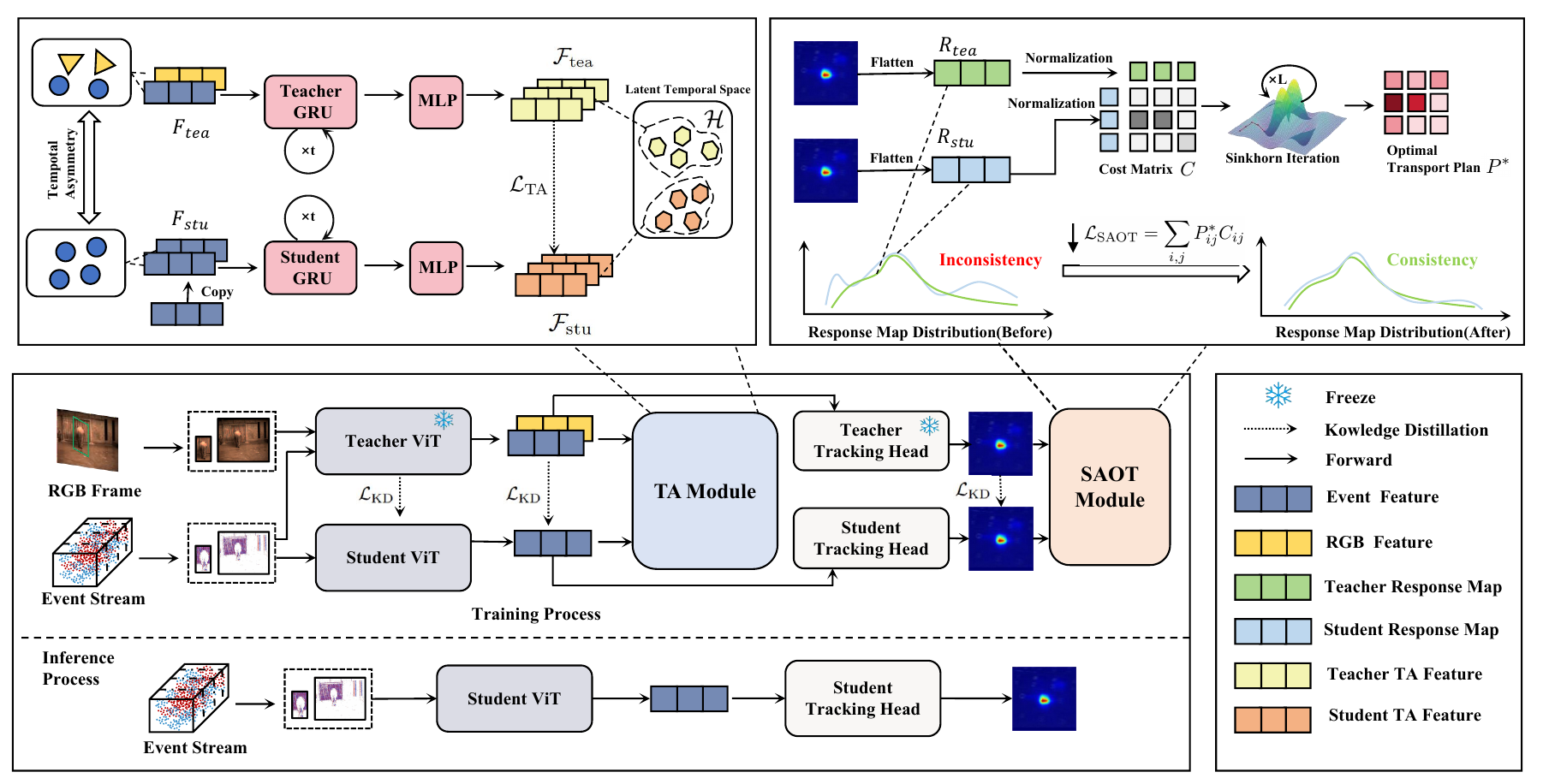}
	\caption{\textbf{Overview of HAD.} A bimodal teacher network (RGB + event) guides an event-only student through {Hierarchical Asymmetric Distillation}. The framework integrates a GRU-based temporal alignment module to synchronize asynchronous feature sequences and an entropic optimal transport-based spatial alignment module to align response distributions across modalities.}
	\label{fig:overview}
\end{figure*}

\subsection{HAD Framework}
\label{sec:had_framework}

As illustrated in \cref{fig:overview}, we propose the Hierarchical Asymmetric Distillation (HAD) framework, which is built upon a Transformer backbone~\cite{vaswani2017attention} and follows a standard knowledge distillation paradigm~\cite{hinton2015distilling}. The RGB sequence $\mathcal{I} = \{I_1, \dots, I_N\}$ and event stream $\mathcal{E} = \{e_1, \dots, e_M\}$ are partitioned into template and search regions, which are then processed by Vision Transformers (ViTs)~\cite{iclr/DosovitskiyB0WZ21}. The teacher network leverages both $\mathcal{I}$ and $\mathcal{E}$ to generate feature representations $F_\mathrm{tea} \in \mathbb{R}^{B \times T_\mathrm{tea} \times L}$, whereas the student network relies solely on $\mathcal{E}$ to produce $F_\mathrm{stu} \in \mathbb{R}^{B \times T_\mathrm{stu} \times L}$.

\subsubsection{Temporal Alignment (TA)}
\label{sec:temporal_hierarchical_alignment}

To bridge the temporal gap between the teacher and student arising from their asynchronous sampling mechanisms, we introduce a Temporal Alignment (TA) module that enforces consistency in their temporal dynamics. Specifically, the teacher processes both RGB and event features $f_\mathrm{RGB} \ll f_\mathrm{event}$, whereas the student observes only high-frequency event streams. Direct frame-wise alignment is infeasible due to the distinct temporal densities and signal characteristics of the two modalities: RGB features are smooth and dense, while event features are sparse and bursty.

To address this, we map both sequences into a latent temporal space where their long-range temporal evolutions can be effectively compared. Each sequence is independently encoded using a Gated Recurrent Unit (GRU)~\cite{corr/ChungGCB14}, which aggregates historical context into a compact temporal representation:
\begin{align}
	h_{k}^t = \mathrm{GRU} \left(F_{k}^t, h_{k}^{t-1} \right), \quad k \in \{\mathrm{tea}, \mathrm{stu}\},
\end{align}
where $F_{k}^t$ denotes the feature at step $t$, and $h_{k}^{t-1}$ represents the previous hidden state. After $T$ iterations, the output $\mathcal{F}_{k} = h_{k}^{T}$ summarizes the temporal dependencies without requiring explicit frame-level supervision.

The final temporal embeddings $h_{k}^{T}$ are projected into a common 768-dimensional latent space through lightweight fully connected layers:
\begin{align}
	\mathcal{F}_\mathrm{k} = \phi_\mathrm{k} \left(h_\mathrm{k}^{T} \right), \quad 
	% \mathcal{F}_\mathrm{stu} = \phi_\mathrm{stu} \left(h_\mathrm{stu}^{T_\mathrm{stu}} \right),
\end{align}
where $\phi(\cdot)$ denotes a projection function.

The TA loss is then defined as the $\ell_2$ distance between these aligned representations:
\begin{align}
	\mathcal{L}_\mathrm{TA} = \left\|\mathcal{F}_\mathrm{stu} - \mathcal{F}_\mathrm{tea} \right\|_2^2.
\end{align}
By minimizing $\mathcal{L}_\mathrm{TA}$, the student is guided to mimic the teacher’s temporal evolution in a rate-agnostic manner, effectively enforcing temporal consistency without requiring frame-level synchronization. This enables robust knowledge transfer across modalities with inherently different temporal structures (see \cref{tab:component_ablation_TA} for ablation results).

\subsubsection{Spatial-Aligned Optimal Transport (SAOT)}
\label{sec:modal_aligned_ot}

In addition to temporal alignment, we enforce spatial consistency between teacher and student response maps. 
Let $R_\mathrm{tea}, R_\mathrm{stu} \in \mathbb{R}^{H \times W}$ denote response maps (with $H = W = 16$ in our implementation). They are normalized using spatial softmax:
\begin{align}
	p_\mathrm{tea} \left(i,j \right) & = \frac{\exp \left(R_\mathrm{tea} \left(i,j \right) \right)}{\sum_{i',j'}\exp \left(R_\mathrm{tea} \left(i',j' \right) \right)}, \\
	p_\mathrm{stu} \left(i,j \right) & = \frac{\exp \left(R_\mathrm{stu} \left(i,j \right) \right)}{\sum_{i',j'}\exp \left(R_\mathrm{stu} \left(i',j' \right) \right)}.
\end{align}
Flattening yields $p = p_\mathrm{tea}^\mathrm{flat}$ and $q = p_\mathrm{stu}^\mathrm{flat} \in \Delta^{HW-1}$. 
The ground-cost matrix $C \in \mathbb{R}^{HW \times HW}$ is defined as:
\begin{align}
	C_{ij} = \left\|x_i - x_j \right\|_2^2.
\end{align}
where $x_i$ and $x_j$ denote 2D pixel coordinates. 
The entropic OT plan solves:
\begin{align}
	P^* = \arg \min_{P \in \Pi \left(p,q \right)} \langle P,C \rangle - \varepsilon \sum_{i,j} P_{ij} \log P_{ij},
\end{align}
where $\Pi(p,q) = \{P \ge 0 \mid P\mathbf{1}_{HW} = p, P^\top\mathbf{1}_{HW} = q\}$, $\langle P,C \rangle = \sum_{i,j} P_{ij}C_{ij}$, and $\varepsilon > 0$ is the regularization strength. 
We solve this problem using Sinkhorn iterations with Gibbs kernel $K_{ij} = \exp(-C_{ij}/\varepsilon)$:
\begin{align}
	\label{eq_uv}
	u^{(l+1)} = \frac{p}{K v^{(l)}}, \quad 
	v^{(l+1)} = \frac{q}{K^\top u^{(l+1)}},
\end{align}
initialized with $v^{(0)} = \mathbf{1}$ and iterated for $l = 0,\dots,L-1$. 
The optimal plan is then:
\begin{align}
	P^* = \mathrm{diag} \left(u^{(L)} \right) K \mathrm{diag} \left(v^{(L)} \right).
\end{align}
This procedure converges linearly to the optimal transport plan $P^*$ for strictly positive cost matrix $C$ and marginals $p,q \in \Delta^{HW-1}$~\cite{sinkhorn1967concerning, nips/Cuturi13}. 
In practice, we fix the number of Sinkhorn iterations to $L = 100$, which ensures stable convergence without incurring significant computational overhead.

The resulting SAOT loss $\mathcal{L}_\mathrm{SAOT}$ is defined as:
\begin{align}
	\mathcal{L}_\mathrm{SAOT} = \sum_{i,j} P^*_{ij} C_{ij}.
\end{align}
Minimizing $\mathcal{L}_\mathrm{SAOT}$ measures the Wasserstein distance between $p_\mathrm{stu}$ and $p_\mathrm{tea}$, explicitly accounting for spatial geometry. 
Unlike $\ell_2$ or Kullback-Leibler (KL) divergence, optimal transport penalizes shifts proportionally to pixel displacement, making it particularly suitable for aligning the student’s sparse event-based responses with the teacher’s dense RGB-based outputs in a geometry-aware manner.

\subsubsection{Optimization Objective}

The total HAD objective combines task, distillation, temporal, and spatial alignment losses:
\begin{align}
	\mathcal{L}_\mathrm{total} = \left(\mathcal{L}_\mathrm{task} + \mathcal{L}_\mathrm{KD} \right) + \lambda_1 \mathcal{L}_\mathrm{TA} + \lambda_2 \mathcal{L}_\mathrm{SAOT},
\end{align}
where $\mathcal{L}_\mathrm{task}$ and $\mathcal{L}_\mathrm{KD}$ follow OSTrack~\cite{eccv/YeCMSC22} and HDETrack~\cite{wang2024event}, and $\lambda_1, \lambda_2$ weight the alignment terms. 
Here, $\mathcal{L}_\mathrm{TA}$ enforces temporal consistency through an $\ell_2$ loss, while $\mathcal{L}_\mathrm{SAOT}$ ensures geometry-aware spatial alignment via the Sinkhorn distance. 
Joint minimization enables the student to inherit task-specific knowledge while aligning temporal dynamics and spatial responses, effectively bridging the gap between RGB and event modalities (see \S\ref{sec:sensitivity_to_alignment_weights} for sensitivity analysis of $\lambda_1$ and $\lambda_2$).

\section{Experimental Results}

\subsection{Datasets and Metrics} 

% We evaluate on three benchmarks:

\textsc{EventVOT}~\cite{wang2024event} comprises 1,141 high-resolution event videos ($1280 \times 720$) spanning 19 object categories and 14 challenging attributes (\textit{e.g.}, low light, fast motion). Its official split includes 841 training, 18 validation, and 282 test sequences captured under diverse conditions (day/night, indoor/outdoor).

\textsc{COESOT}~\cite{tang2022coesot} contains 1,354 RGB-Event bimodal sequences ($346 \times 260$) across 90 categories in highly dynamic scenes, with 827 sequences for training and 527 for testing. Each frame is densely annotated with an absence flag and 17 attributes (\textit{e.g.}, occlusion, low light).

\textsc{VisEvent}~\cite{wang2021viseventbenchmark} includes 820 RGB-Event video pairs ($346 \times 260$) spanning 17 object categories and 17 challenging attributes (\textit{e.g.}, low illumination, fast motion, motion blur, background clutter). The official split contains 500 training and 320 test sequences, totaling 371,128 densely annotated frames.

Tracking performance is evaluated using three metrics: Success Rate (SR), Precision Rate (PR)~\cite{cvpr/WuLY13}, and Normalized Precision Rate (NPR)~\cite{eccv/MullerBGAG18}. SR measures the percentage of frames whose predicted bounding boxes overlap sufficiently with the ground truth. PR measures the proportion of frames in which the predicted center lies within a distance threshold of the ground truth. NPR normalizes PR for scale-invariant comparison. Together, these complementary metrics comprehensively capture tracking accuracy, precision, and robustness.

\subsection{Implementation Details}

To ensure fair and reproducible comparisons, all training configurations, including batch size, learning-rate schedule, optimizer settings, and data augmentations, are aligned with the official implementation of HDETrack~\cite{wang2024event}.

All models are implemented in PyTorch~\cite{paszke2019pytorch} and trained on NVIDIA RTX~3090 GPU using AdamW~\cite{loshchilov2017decoupled} with an initial learning rate of $4 \times 10^{-4}$, weight decay of $1 \times 10^{-4}$, and batch size of 38 for the pure-event dataset \textsc{EventVOT}~\cite{wang2024event} and 32 for the RGB-Event datasets \textsc{COESOT}~\cite{tang2022coesot} and \textsc{VisEvent}~\cite{wang2021viseventbenchmark}. Training runs for 50~epochs with learning-rate decay at epoch~40. Data augmentations include bounding-box jittering, search-area cropping, normalization, and random flipping. Inference FPS is measured on a single RTX~3090 GPU.

For single-modal datasets such as \textsc{EventVOT}, the event stream is converted into both event voxels and event frames: the student consumes voxels, while the teacher uses both frames and voxels. For RGB-Event datasets, the student receives event frames, and the teacher receives both RGB and event frames. This design allows the student to be distilled from richer multimodal cues while remaining event-only at inference.

\begin{table*}[!t]
	\centering
	\setlength{\tabcolsep}{6pt}
	\caption{\textbf{Comparison of state-of-the-art trackers on \textsc{EventVOT}, \textsc{COESOT}, and \textsc{VisEvent}.} 
	$\dagger$ denotes reproduced results. Values in \textbf{bold} and \underline{underline} indicate the best and second-best results, respectively.}
	\begin{tabular}{c|ll|cc|ccc|ccc|ccc}
	\toprule[1.1pt]
	\multirow{2}[2]{*}{Type} & \multirow{2}[2]{*}{Tracker} & \multirow{2}[2]{*}{Venue} & \multirow{2}[2]{*}{FPS} & \multirow{2}[2]{*}{\makecell{Params \\ (M)}} & 
	\multicolumn{3}{c|}{\textsc{EventVOT}} & 
	\multicolumn{3}{c|}{\textsc{COESOT}} & 
	\multicolumn{3}{c}{\textsc{VisEvent}} \\
	\cmidrule(lr){6-8} \cmidrule(lr){9-11} \cmidrule(lr){12-14}
	& & & & & SR & PR & NPR & SR & PR & NPR & SR & PR & NPR \\
	\midrule
	\multirow{7}{*}{RGB-E} 
	& TrDiMP~\cite{cvpr/0020ZWL21} & CVPR’21 & 26 & 26.3 & 39.9 & 35.3 & 47.2 & 50.7 & 56.9 & 55.2 & 60.1 & 72.2 & 71.7 \\
	& ToMP50~\cite{cvpr/0007DBPPYG22} & CVPR’22 & 25 & 26.1 & 37.6 & 33.5 & 45.6 & 46.3 & 52.9 & 52.5 & 59.8 & 70.8 & 70.9 \\
	& CEUTrack~\cite{tang2022coesot} & arXiv’22 & 75 & - & - & - & - & 62.7 & 76.0 & 74.9 & 64.9 & 69.0 & 73.8 \\
	& HRCEUTrack~\cite{iccv/ZhuHW23} & ICCV’23 & - & - & - & - & - & 63.2 & 71.9 & 70.2 & - & - & - \\
	& CSAM-T~\cite{nips/ZhangD0DH24} & NeurIPS’24 & - & - & - & - & - & 63.3 & 73.3 & 70.5 & 61.5 & 76.1 & 72.4 \\
	& CSAM-B~\cite{nips/ZhangD0DH24} & NeurIPS’24 & 53 & 106.9 & - & - & - & 68.1 & 76.7 & 74.8 & 65.9 & 81.6 & 78.6 \\
	& Cross-EI~\cite{tip/ChenWDLS25} & TIP’25 & - & 16.7 & - & - & - & 61.7 & 70.9 & - & 53.1 & 93.0 & - \\
	\midrule
	\multirow{7}{*}{Event}
	& STARK~\cite{iccv/0002PF0L21} & ICCV’21 & 42 & 28.1 & 44.5 & 39.6 & 52.0 & 40.8 & 44.9 & 44.4 & 34.8 & 41.8 & - \\
	& TransT~\cite{cvpr/CuiJ0W22} & CVPR’21 & 50 & 18.0 & 54.3 & 53.5 & 63.2 & 45.6 & 51.4 & 50.4 & \textbf{39.5} & 47.1 & - \\
	% SimTrack~\cite{eccv/ChenLBQSLGWO22} & Event & ECCV’22 & 40 & 57.8 & 55.4 & 54.2 & 64.1 & 48.3 & 53.5 & 52.9 & - & - & - \\
	& OSTrack~\cite{eccv/YeCMSC22} & ECCV’22 & 105 & 92.1 & 55.4 & 56.4 & 65.2 & 50.9 & 57.8 & 56.7 & 34.5 & 48.9 & 38.5 \\
	& AiATrack~\cite{eccv/GaoZMWY22} & ECCV’22 & 38 & 15.8 & \underline{57.4} & 56.4 & 66.7 & 51.3 & 57.9 & 56.2 & - & - & - \\
	& HDETrack~\cite{wang2024event} ${\dagger}$ & CVPR’24 & 107 & 92.1 & 56.5 & 56.5 & 65.8 & \underline{52.6} & \underline{59.6} & 58.5 & 36.1 & \underline{51.3} & \underline{39.7} \\
	& SFTrack-Fast~\cite{corr/abs-2505-12903} & arXiv’25 & 126 & 50.4 & 53.8 & 55.0 & \textbf{69.0} & 49.3 & 59.1 & \textbf{59.8} & - & - & - \\
	\rowcolor{gray!20} & HAD (Ours) & & 107 & 92.1 & \textbf{57.8} & \textbf{58.0} & \underline{67.2} & \textbf{52.9} & \textbf{60.0} & \underline{58.8} & \underline{36.7} & \textbf{51.8} & \textbf{40.0} \\
	\bottomrule[1.1pt]
	\end{tabular}
	\label{tab:comparison}
\end{table*}

In \cref{tab:comparison}, all results except our reproduced HDETrack and HAD are cited directly from their respective papers. Metrics for TrDiMP~\cite{cvpr/0020ZWL21}, ToMP50~\cite{cvpr/0007DBPPYG22}, STARK~\cite{iccv/0002PF0L21}, TransT~\cite{cvpr/CuiJ0W22}, OSTrack~\cite{eccv/YeCMSC22}, AiATrack~\cite{eccv/GaoZMWY22}, and HDETrack~\cite{wang2024event} follow \cite{wang2024event}; all other trackers are referenced from their original sources.

\subsection{Comparison with State-of-the-Art}

We compare HAD with state-of-the-art trackers under two settings: RGB-Event bimodal input and event-only input. In general, bimodal trackers achieve higher accuracy by directly leveraging RGB information but often at the cost of real-time efficiency. For fair evaluation, our main analysis focuses on unimodal (event-only) comparisons, while bimodal results are provided for completeness.

\begin{figure}[!t]
	\centering
	\includegraphics[width = 0.8\linewidth]{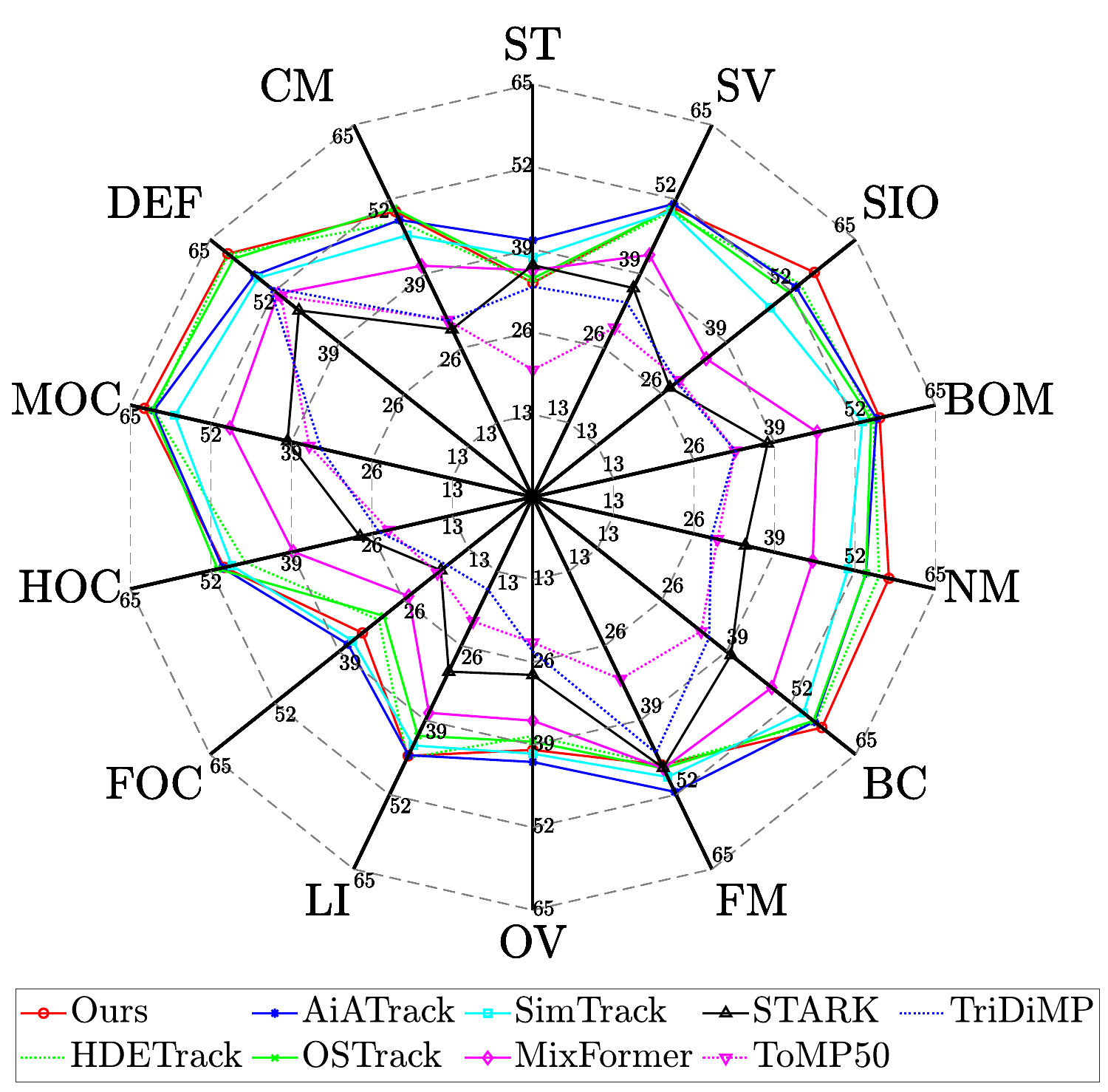}
	\caption{\textbf{Radar charts of PR metrics on \textsc{EventVOT}.} Each axis corresponds to a specific tracking challenge, illustrating performance across different attributes. Zooming in reveals finer performance differences among competing methods.}
	\label{fig:radar_eventvot}
\end{figure}

\subsubsection{Results on \textsc{EventVOT}}

As shown in \cref{tab:comparison}, HAD sets a new state of the art with SR~57.8\%, PR~58.0\%, and NPR~67.2\%. In per-attribute evaluation (see \cref{fig:radar_eventvot}), HAD performs best under low illumination (LI), deformation (DEF), no motion (NM), and background object motion (BOM). These gains directly tackle key event-based tracking challenges, sparse static scenes and low-light robustness.

\subsubsection{Results on \textsc{COESOT}}

On \textsc{COESOT}, HAD achieves SR~52.9\%, PR~60.0\%, and NPR~58.8\%, outperforming HDETrack~\cite{wang2024event} by +1.3\%, +1.5\%, and +1.4\%, respectively. These improvements demonstrate the effective integration of RGB texture cues with event-stream motion information: RGB offers stable spatial detail, while events enhance dynamic perception in high-speed and low-light conditions. Importantly, HAD achieves these gains without increasing parameters (92.1M) or sacrificing speed (107~FPS), confirming an excellent balance between efficiency and accuracy.

\begin{table}[!t]
	\centering
	\setlength{\tabcolsep}{1.8pt}
	\caption{\textbf{Training performance comparison between HDETrack~\cite{wang2024event} and HAD on \textsc{EventVOT} and \textsc{COESOT}.} GPU utilization (\%) is reported as a dynamic range (min-max), training duration in hours (h), and memory usage in gigabytes (GB).}
	\begin{tabular}{l|c|cc|cc}
	\toprule[1.1pt]
	\multirow{2}[2]{*}{Method} & \multirow{2}[2]{*}{GPU} & \multicolumn{2}{c|}{Training Duration} & \multicolumn{2}{c}{Memory Usage} \\
	\cmidrule(lr){3-4} \cmidrule(lr){5-6}
	& & \textsc{EventVOT} & \textsc{COESOT} & \textsc{EventVOT} & \textsc{COESOT} \\
	\midrule
	HDETrack & 83-100 & 5.25 & 5.11 & 15.74 & 13.52 \\
	\rowcolor{gray!20}
	HAD & 85-100 & 5.65 & 5.83 & 16.84 & 14.52 \\
	\bottomrule[1.1pt]
	\end{tabular}
	\label{tab:efficient}
\end{table}

\begin{figure*}[!t]
	\centering
	\tabcolsep = 1pt
	\begin{tabular}{cccc}
	\cellcolor{white}{} & \footnotesize \textbf{Col 1} & \footnotesize \textbf{Col 2} & \footnotesize \textbf{Col 3} \\[-2pt]

	\rotatebox{90}{\parbox{4cm}{\centering \footnotesize \textbf{Row 1}}} &
	\includegraphics[width = 0.3\linewidth, frame]{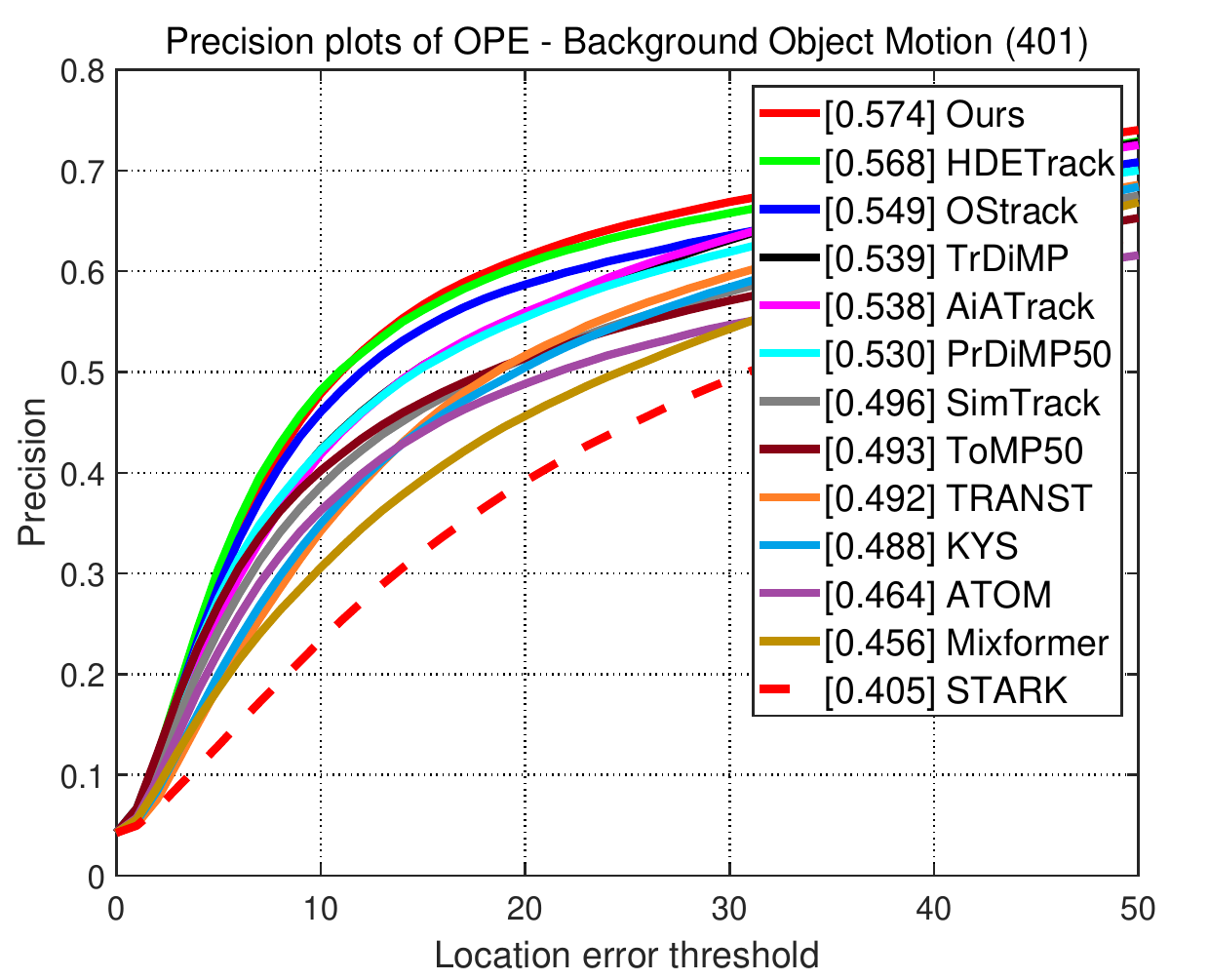} &
	\includegraphics[width = 0.3\linewidth, frame]{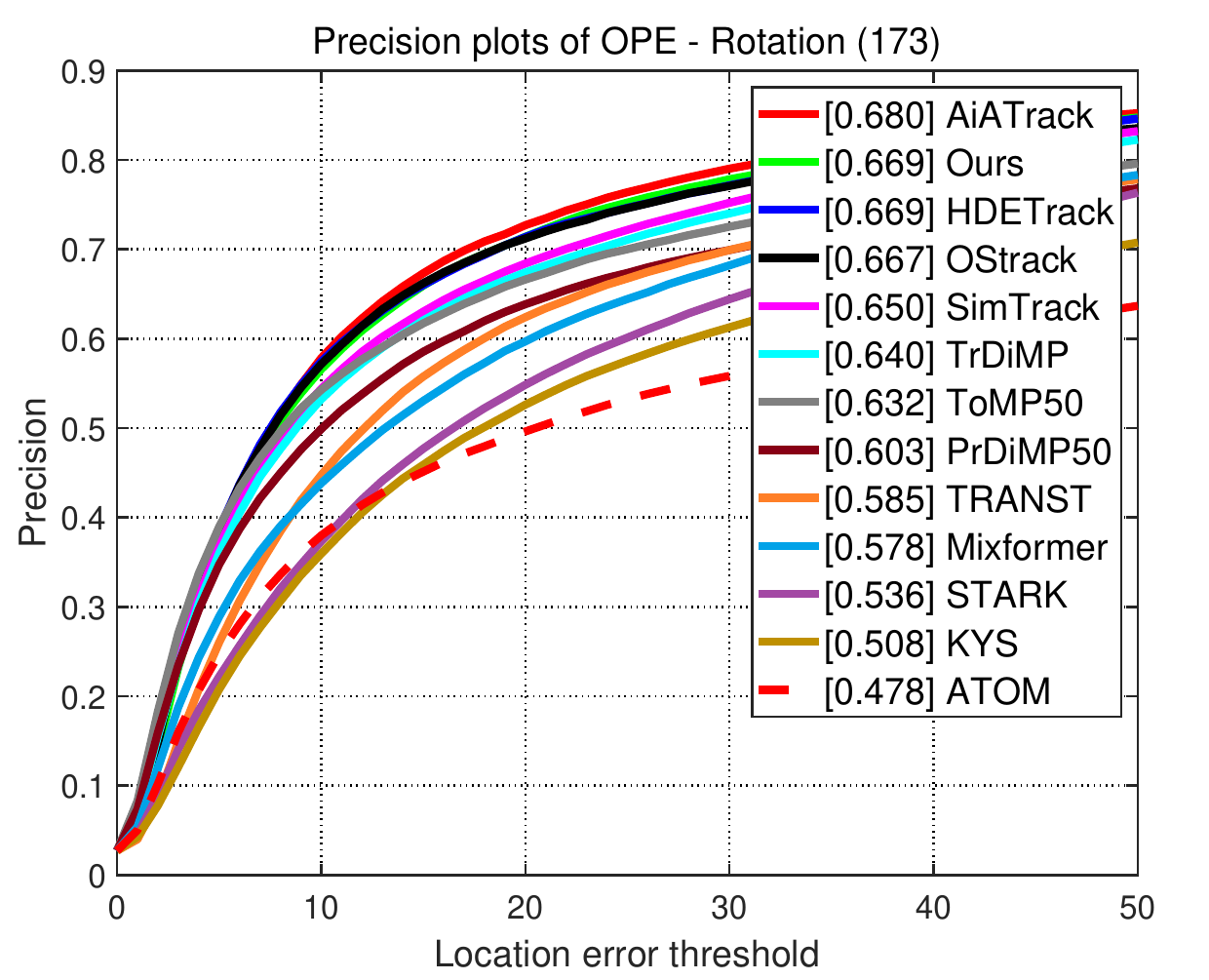} &
	\includegraphics[width = 0.3\linewidth, frame]{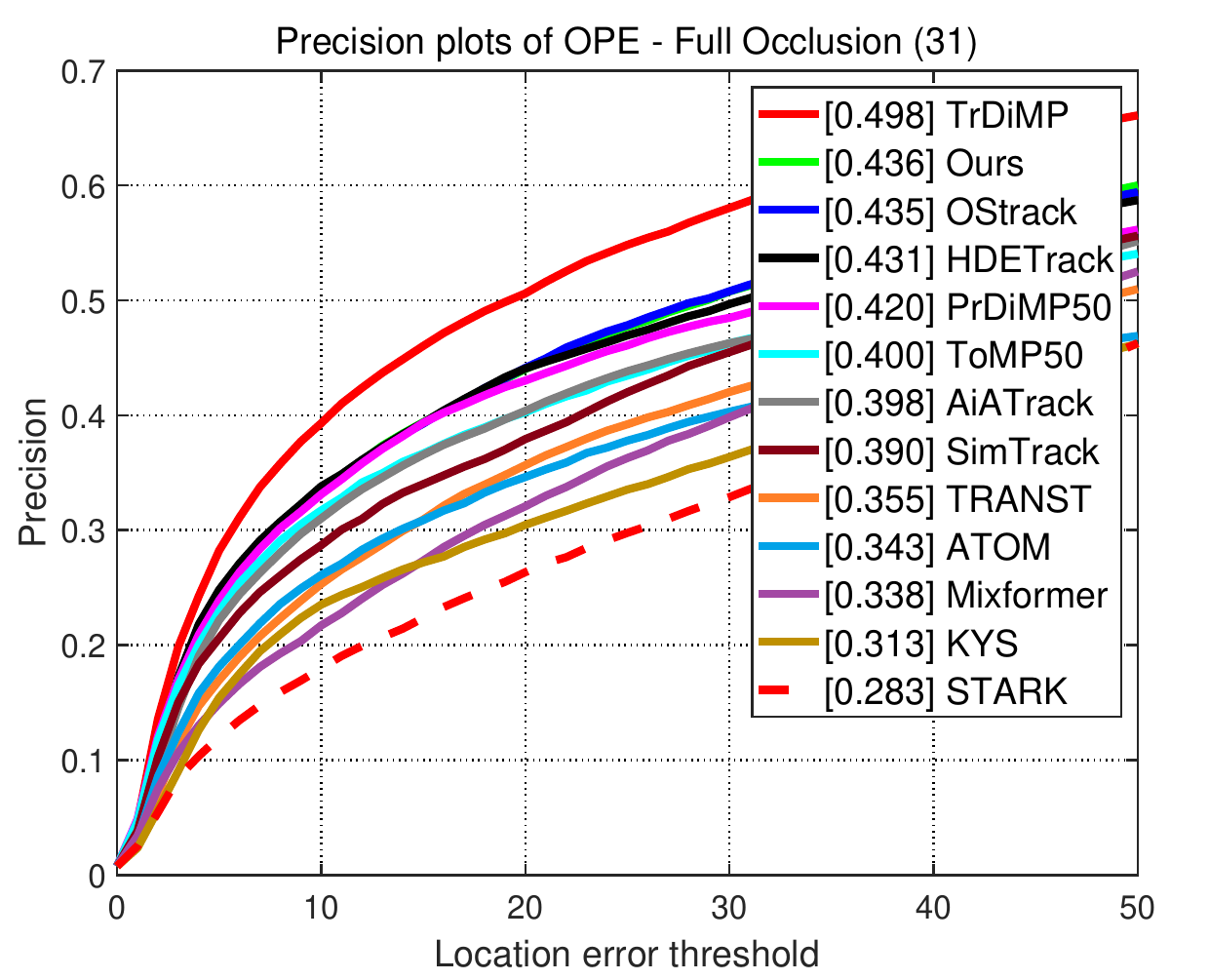} \\[-1pt]

	\rotatebox{90}{\parbox{4cm}{\centering \footnotesize \textbf{Row 2}}} &
	\includegraphics[width = 0.3\linewidth, frame]{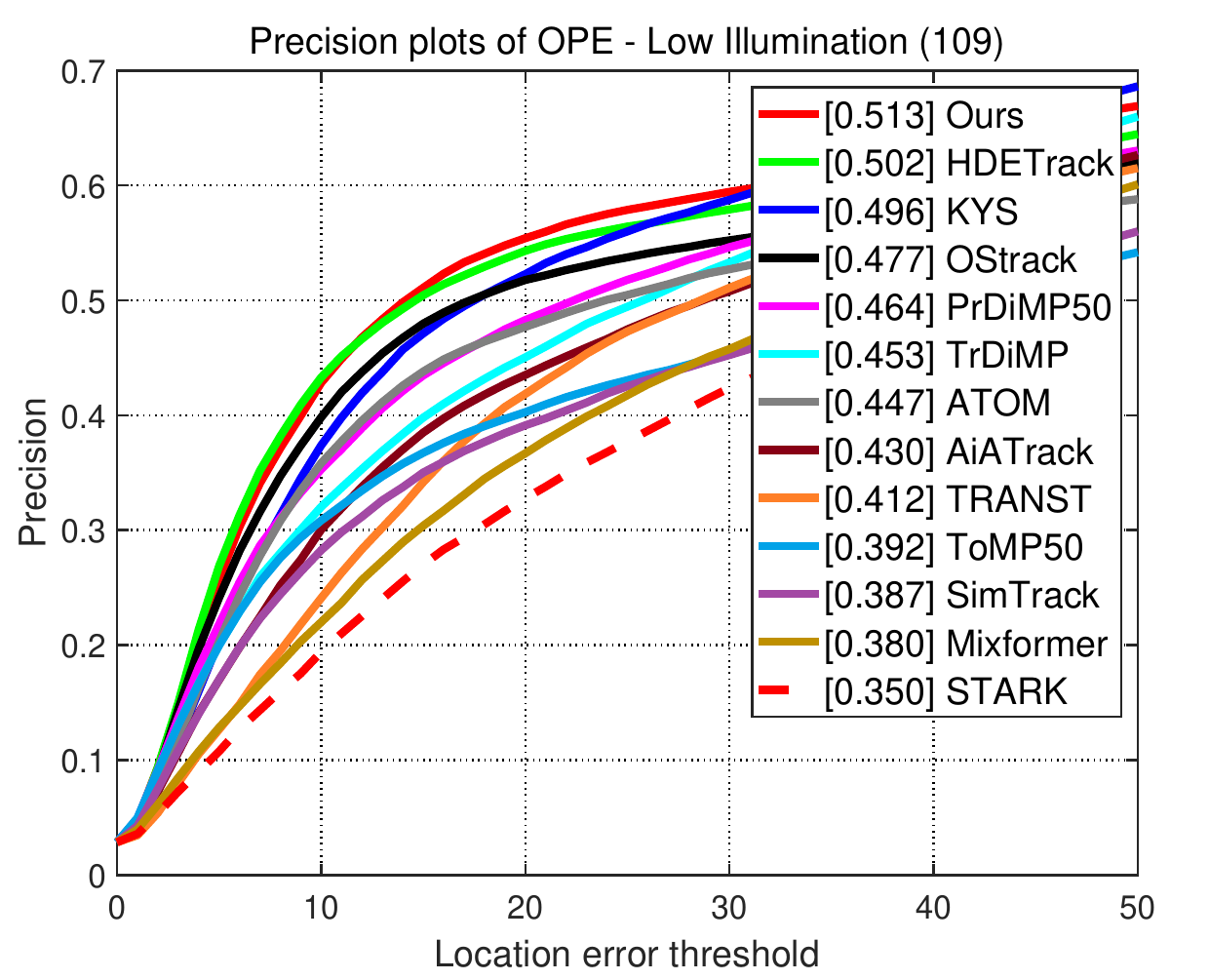} &
	\includegraphics[width = 0.3\linewidth, frame]{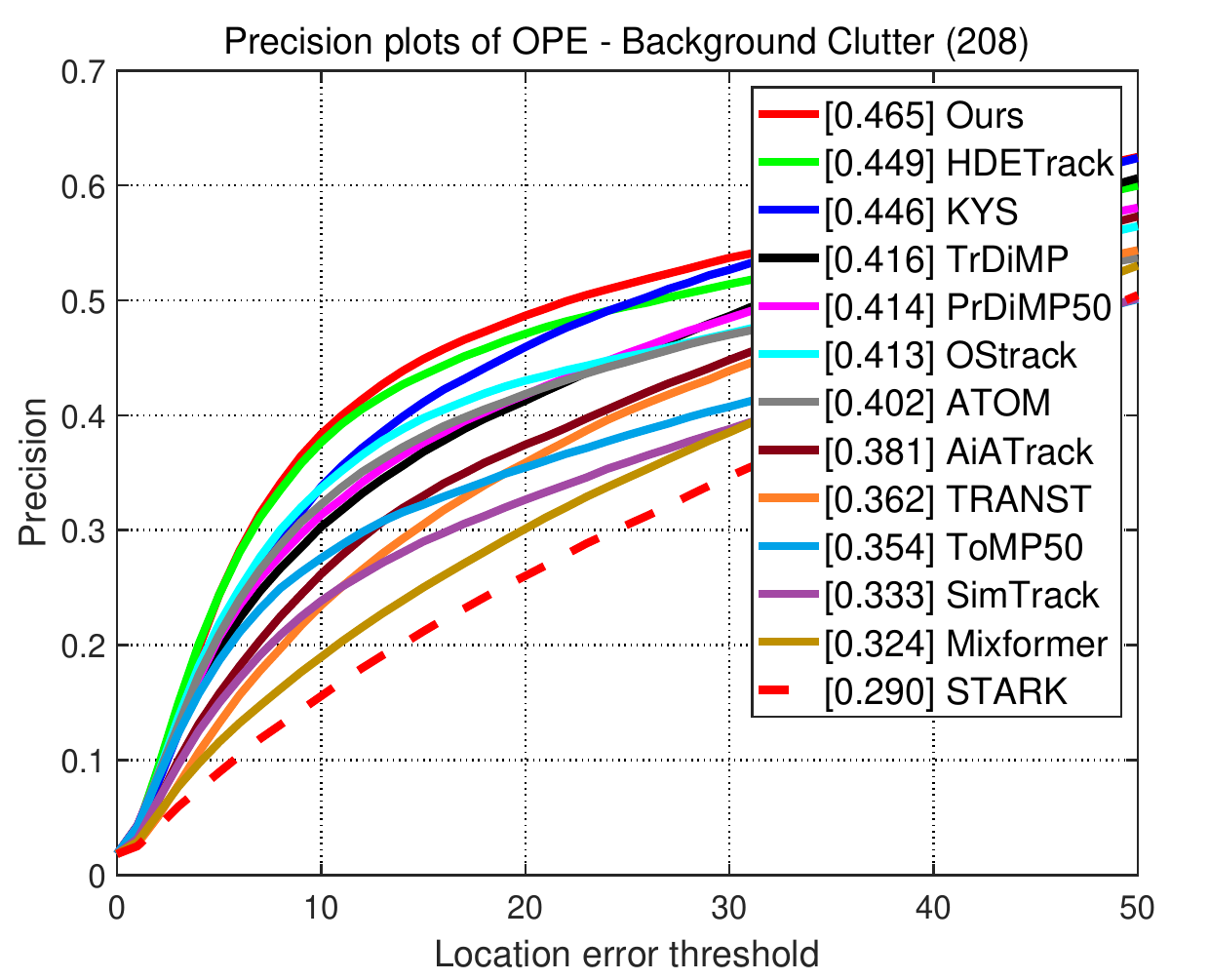} &
	\includegraphics[width = 0.3\linewidth, frame]{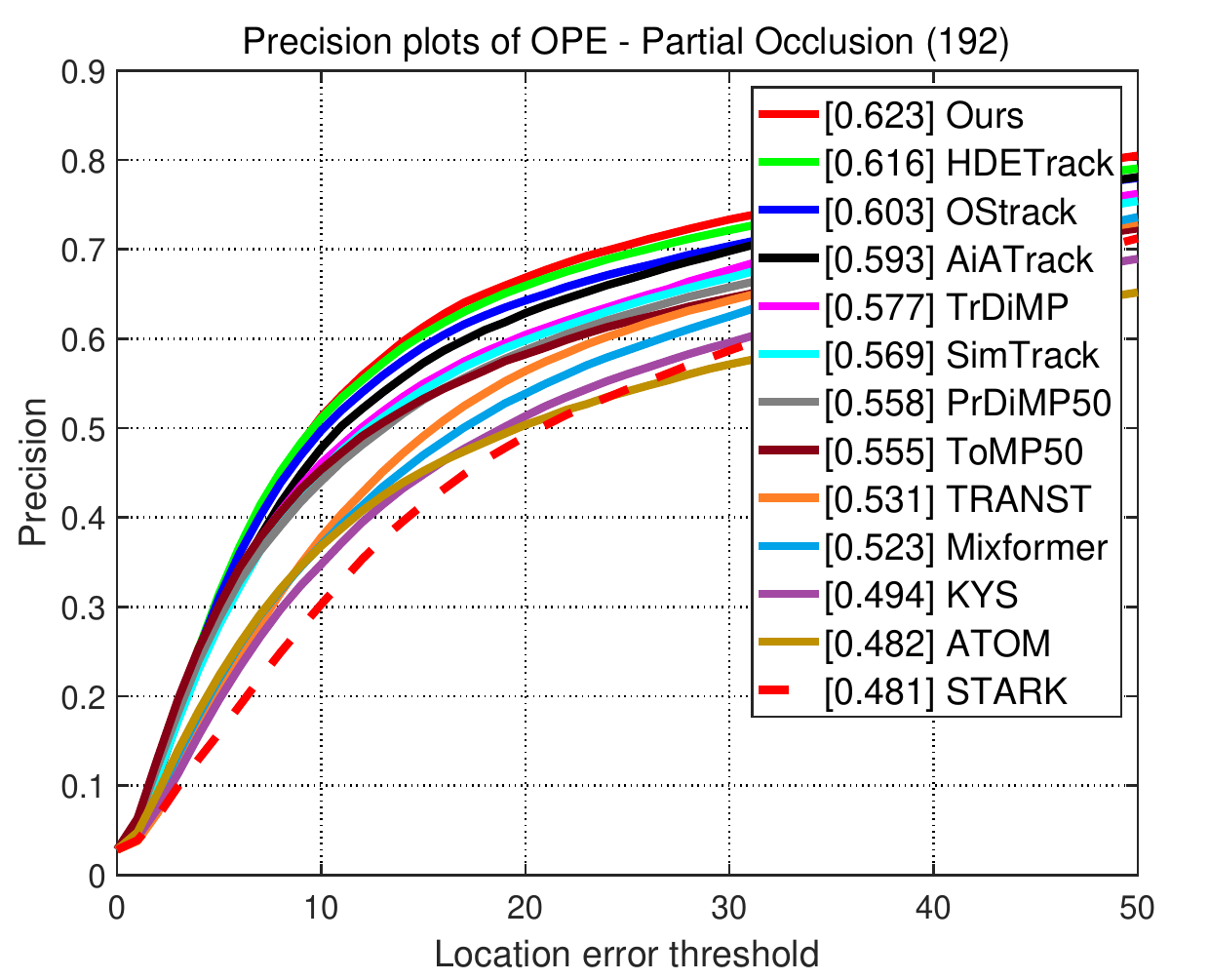} \\[-1pt]

	\rotatebox{90}{\parbox{4cm}{\centering \footnotesize \textbf{Row 3}}} &
	\includegraphics[width = 0.3\linewidth, frame]{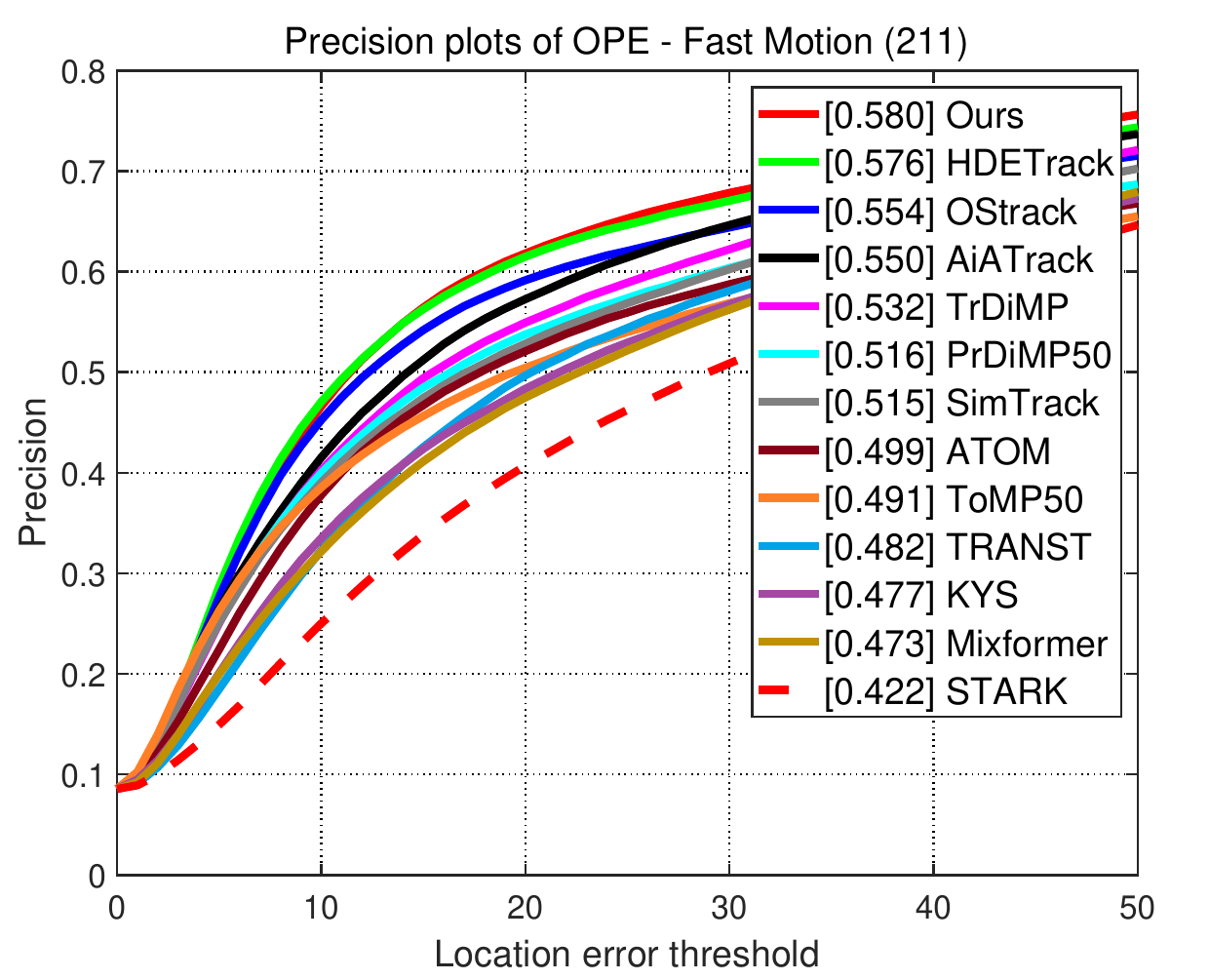} &
	\includegraphics[width = 0.3\linewidth, frame]{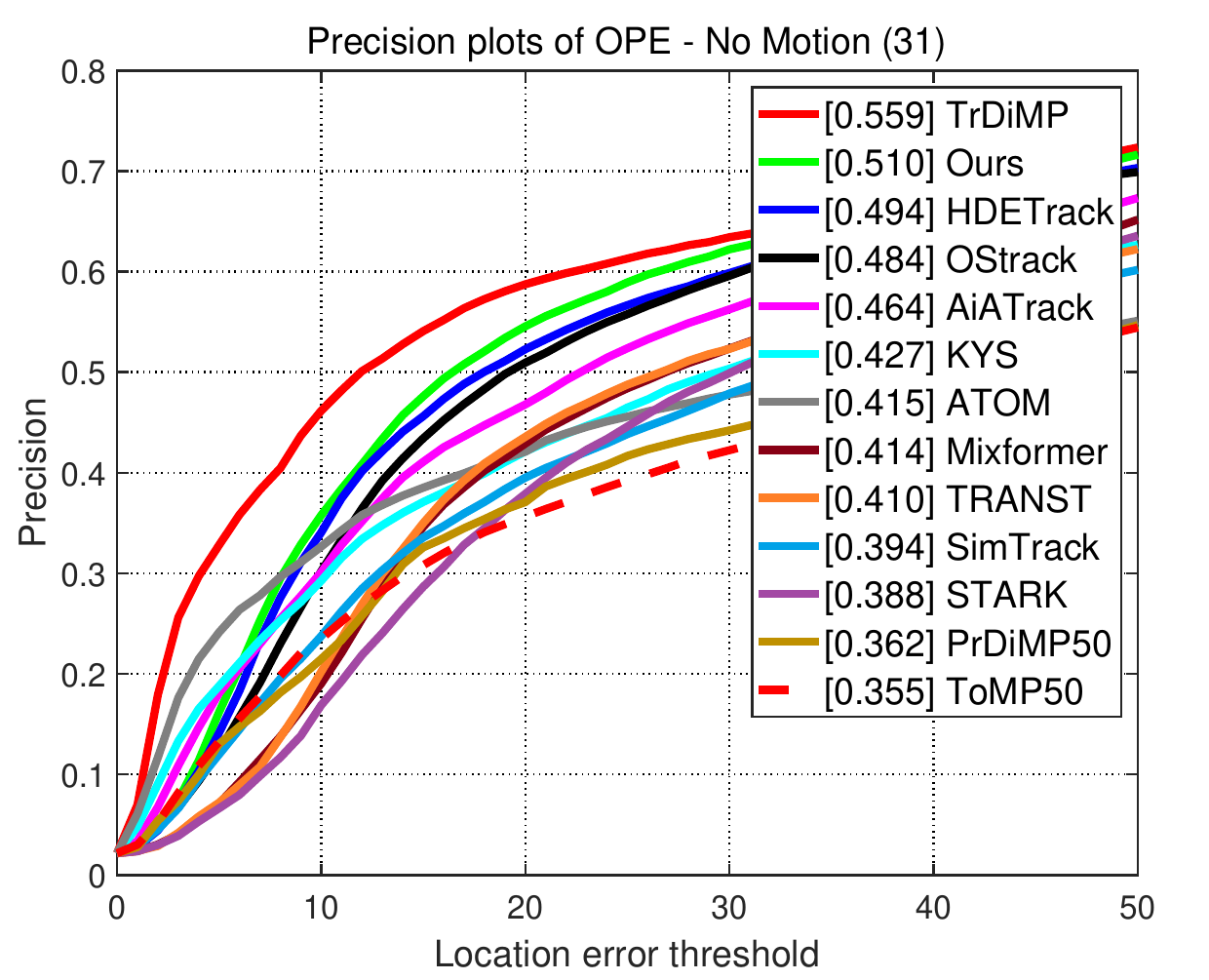} &
	\includegraphics[width = 0.3\linewidth, frame]{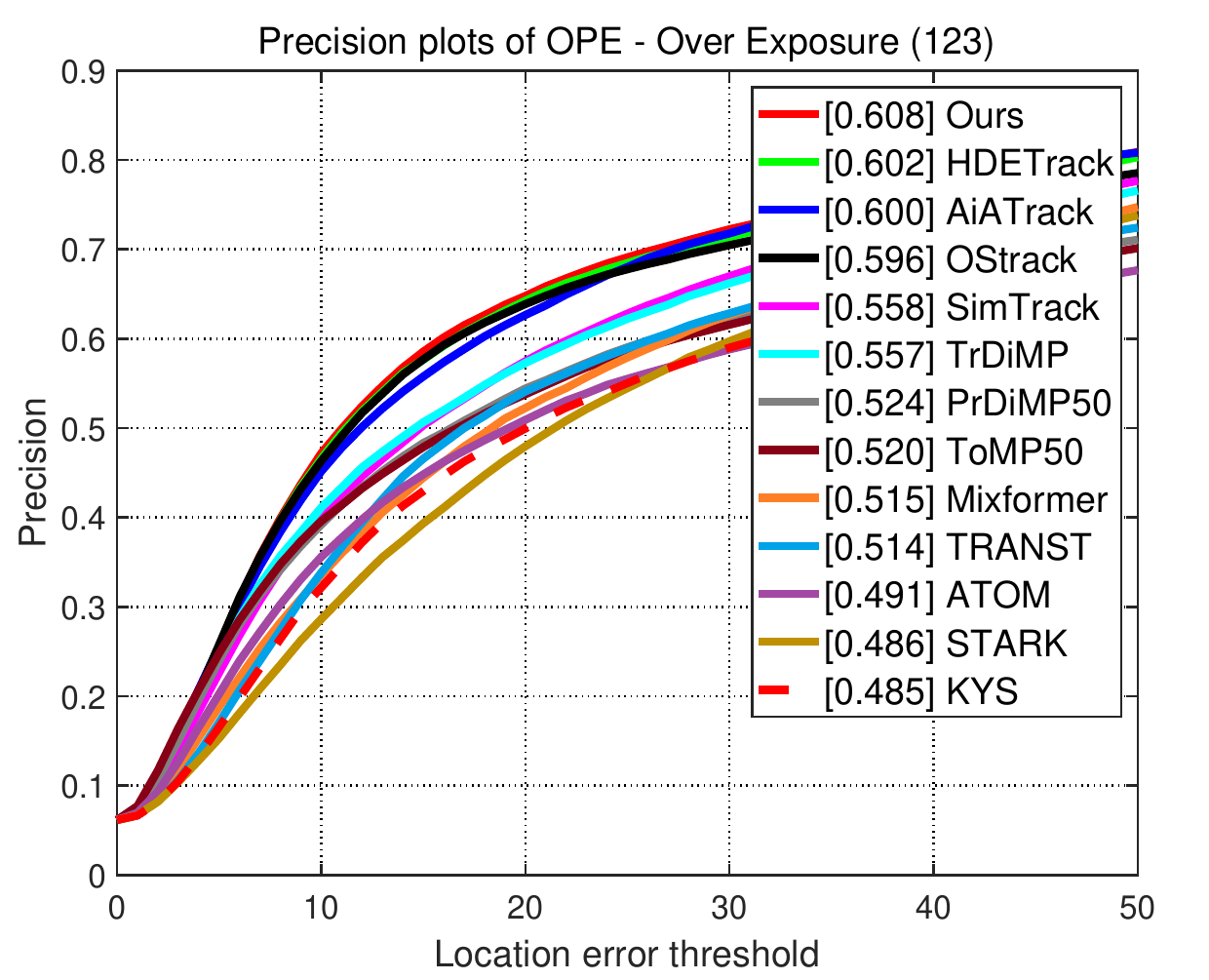} \\
	\end{tabular}
	\caption{\textbf{Comparison of precision rates for challenging sequences on \textsc{COESOT}.} Each subgraph title corresponds to a specific challenge, with the number in parentheses indicating the number of video sequences. Zooming in reveals finer performance differences among competing methods.}
	\label{fig:coesot_challenge_pr}
\end{figure*}

\begin{table*}[!t]
	\centering
	\setlength{\tabcolsep}{10pt}
	\caption{\textbf{Component stacking ablation on \textsc{EventVOT}, \textsc{COESOT}, and \textsc{VisEvent}.} \textit{Base} denotes the unimodal student baseline trained using the HDETrack~\cite{wang2024event} distillation strategy.}
	\begin{tabular}{c|ccc|ccc|ccc|ccc}
	\toprule[1.1pt]
	\multirow{2}[2]{*}{No.} & \multirow{2}[2]{*}{Base} & \multirow{2}[2]{*}{TA} & \multirow{2}[2]{*}{SAOT} & 
	\multicolumn{3}{c|}{\textsc{EventVOT}} & 
	\multicolumn{3}{c|}{\textsc{COESOT}} &
	\multicolumn{3}{c}{\textsc{VisEvent}} \\
	\cmidrule(lr){5-7} \cmidrule(lr){8-10} \cmidrule(lr){11-13}
	& & & & SR & PR & NPR & SR & PR & NPR & SR & PR & NPR \\
	\midrule
	1 & \CIRCLE & \Circle & \Circle & 56.5 & 56.5 & 65.8 & 52.6 & 59.6 & 58.5 & 36.1 & 51.3 & 39.7 \\
	2 & \CIRCLE & \CIRCLE & \Circle & 57.4 & 57.7 & 66.7 & 52.8 & 59.9 & 58.7 & 36.3 & 51.6 & 39.9 \\
	3 & \CIRCLE & \Circle & \CIRCLE & 57.7 & 57.7 & 66.0 & 52.7 & 59.8 & 58.5 & 36.2 & 51.4 & 39.9 \\
	\rowcolor{gray!20}
	4 & \CIRCLE & \CIRCLE & \CIRCLE & \textbf{57.8} & \textbf{58.0} & \textbf{67.2} & \textbf{52.9} & \textbf{60.0} & \textbf{58.8} & \textbf{36.7} & \textbf{51.8} & \textbf{40.0} \\
	\bottomrule[1.1pt]
	\end{tabular}
	\label{tab:component}
\end{table*}

\cref{fig:coesot_challenge_pr} shows PR across \textsc{COESOT} attributes. HAD attains the highest PR in BOM (R1C1), LI (R2C1), BC (R2C2), PO (R2C3), FM (R3C1), and OE (R4C4). TrDiMP~\cite{cvpr/0020ZWL21} slightly surpasses HAD in Full Occlusion (FO, R1C3) and No Motion (NM, R3C2) due to direct RGB access and advanced propagation. Nevertheless, HAD, restricted to event-only inference, still narrows the gap and surpasses several bimodal methods, validating the effectiveness of asymmetric distillation in transferring essential RGB knowledge to the event domain.

\subsubsection{Results on \textsc{VisEvent}}

On \textsc{VisEvent}, HAD achieves SR~36.7\%, PR~51.8\%, and NPR~40.0\%, outperforming the previous best unimodal tracker HDETrack~\cite{wang2024event} by +0.6\%, +0.5\%, and +0.3\%, respectively. While RGB-Event trackers perform markedly better on \textsc{VisEvent} than on \textsc{COESOT}, event-only trackers show the opposite trend. This suggests that \textsc{VisEvent} relies more heavily on RGB content. Consequently, HAD’s relative improvement over HDETrack is more significant on \textsc{VisEvent}, confirming its capability to bridge spatial-temporal asymmetry even where RGB information dominates.

\subsubsection{Comparison with SFTrack-Fast}
As shown in \cref{tab:comparison}, HAD surpasses SFTrack-Fast~\cite{corr/abs-2505-12903} in SR (+4.0\%) and PR (+3.0\%) on \textsc{EventVOT}, and in SR (+3.6\%) and PR (+0.9\%) on \textsc{COESOT}. SFTrack-Fast exhibits marginally higher NPR.

Since the NPR metric is more sensitive to small targets and scale variations~\cite{eccv/MullerBGAG18}, this indicates that SFTrack-Fast can respond more rapidly to sequences involving small objects or significant target-scale changes. Its low-latency characteristic effectively reduces normalized localization errors, thereby improving NPR performance. In contrast, HAD focuses more on maintaining overall spatio-temporal consistency, achieving superior results in absolute localization accuracy and tracking success rate.

\subsection{Efficiency Analysis}

\cref{tab:efficient} compares HAD with HDETrack~\cite{wang2024event} on \textsc{EventVOT} and \textsc{COESOT}. HAD requires slightly longer training (5.65~h \textit{vs.} 5.25~h; 5.83~h \textit{vs.} 5.11~h) and modestly higher memory (16.84~GB \textit{vs.} 15.74~GB; 14.52~GB \textit{vs.} 13.52~GB), but achieves higher GPU utilization (85-100\% \textit{vs.} 83-100\%), indicating better resource usage. The additional memory cost remains below 1.1~GB, and the slight increase in training time is outweighed by consistent accuracy gains. 

Overall, considering both the FPS indicators in \cref{tab:comparison} and the efficiency comparisons in \cref{tab:efficient}, HAD achieves a favorable balance between performance and computational cost, making it well-suited for scenarios where accuracy is prioritized over marginal resource savings.

\begin{table}[!t]
	\centering
	\setlength{\tabcolsep}{6pt}
	\caption{\textbf{Ablation of TA implementations.} \texttt{TA Implementation} indicates the network architecture adopted for the temporal alignment module.}
	\begin{tabular}{c|ccc|ccc}
	\toprule[1.1pt]
	\multirow{2}[2]{*}{\makecell{TA \\ Implementation}} & 
	\multicolumn{3}{c|}{\textsc{EventVOT}} & 
	\multicolumn{3}{c}{\textsc{COESOT}} \\
	\cmidrule(lr){2-4} \cmidrule(lr){5-7}
	& SR & PR & NPR & SR & PR & NPR \\
	\midrule
	RNN & 57.6 & 57.8 & 67.0 & 52.4 & 59.6 & 58.4 \\
	MLP & 56.7 & 56.9 & 65.8 & 51.8 & 58.5 & 57.4 \\
	Mamba & 56.9 & 57.0 & 66.3 & 52.4 & 59.5 & 58.3 \\
	Bi-GRU & 56.8 & 57.0 & 66.1 & 51.9 & 58.7 & 57.8 \\
	Bi-LSTM & 56.9 & 57.3 & 66.3 & 52.1 & 59.0 & 57.9 \\
	\rowcolor{gray!20}
	GRU & \textbf{57.8} & \textbf{58.0} & \textbf{67.2} & \textbf{52.9} & \textbf{60.0} & \textbf{58.8} \\
	\bottomrule[1.1pt]
	\end{tabular}
	\label{tab:component_ablation_TA}
\end{table}

\begin{figure}[!t]
\centering
\pgfplotsset{
 ablation plot/.style = {
 width = 4.8cm,
 height = 4.0cm,
 enlarge x limits = 0.15, 
 ymin = 50, ymax = 70,
 title style = {font = \footnotesize},
 tick label style = {font = \scriptsize},
 grid = major,
 grid style = {dashed,gray!30},
 legend style = {
 font = \footnotesize,
 legend columns = -1, 
 at = {(0.5,-0.2)}, 
 anchor = north, 
 /tikz/every even column/.append style = {column sep = 1em},
 }
 }
}
% ---------------------------------------------
% Left: TA Module
% ---------------------------------------------
\begin{minipage}{0.48\linewidth}
\centering
\begin{tikzpicture}
\begin{axis}[
 ablation plot,
 symbolic x coords = {1,3,5,7,10,12},
 xtick = data,
 width = 5.2cm,
 title = {TA Module}
]
\addplot[
orange, mark = *, 
nodes near coords = {\pgfmathprintnumber[fixed, precision = 1]{\pgfplotspointmeta}}, 
every node near coord/.style = {font = \tiny, anchor = north}] coordinates {(1,57.2)(3,56.5)(5,56.7)(7,57.3)(10,57.4)(12,57.3)};
\addplot[
blue, 
mark = pentagon*, 
nodes near coords = {\pgfmathprintnumber[fixed, precision = 1]{\pgfplotspointmeta}}, 
every node near coord/.style = {font = \tiny, anchor = south}] coordinates {(1,57.3)(3,56.8)(5,57.0)(7,57.4)(10,57.7)(12,57.6)};
\addplot[
green, 
mark = triangle*, 
nodes near coords = {\pgfmathprintnumber[fixed, precision = 1]{\pgfplotspointmeta}}, 
every node near coord/.style = {font = \tiny, anchor = south}] coordinates {(1,66.6)(3,65.9)(5,66.0)(7,66.7)(10,66.7)(12,66.7)};
\end{axis}
\end{tikzpicture}
\end{minipage}
% \vfill
\hfill
% ---------------------------------------------
% Right: SAOT Module
% ---------------------------------------------
\begin{minipage}{0.49\linewidth}
\centering
\begin{tikzpicture}
\begin{axis}[
 ablation plot,
 symbolic x coords = {1,3,5,7,10},
 xtick = data,
 yticklabels = \empty
 width = 5.2cm,
 title = {SAOT Module}
]
\addplot[
orange, mark = *, 
nodes near coords = {\pgfmathprintnumber[fixed, precision = 1]{\pgfplotspointmeta}}, 
every node near coord/.style = {font = \tiny, anchor = north}] coordinates {(1,57.3)(3,57.7)(5,56.6)(7,56.2)(10,56.2)};
\addplot[
blue, 
mark = pentagon*, 
nodes near coords = {\pgfmathprintnumber[fixed, precision = 1]{\pgfplotspointmeta}}, 
every node near coord/.style = {font = \tiny, anchor = south}] coordinates {(1,57.6)(3,57.7)(5,56.6)(7,56.5)(10,56.4)};
\addplot[
green, 
mark = triangle*, 
nodes near coords = {\pgfmathprintnumber[fixed, precision = 1, fixed zerofill]{\pgfplotspointmeta}}, 
every node near coord/.style = {font = \tiny, anchor = south}] coordinates {(1,66.6)(3,67.0)(5,65.9)(7,65.8)(10,65.6)};
\end{axis}
\end{tikzpicture}
\end{minipage}
\begin{tikzpicture}[baseline = (current bounding box.center)]
 \def\dx{1.5}
 \draw[orange, thick] (0,0.15) -- (0.3,0.15);
 \node[orange, mark = *, mark size = 2pt] at (0.25,0) {};
 \node[anchor = west, font = \footnotesize\bfseries] at (0.4,0.15) {SR};
 \draw[blue, thick] (\dx,0.15) -- (\dx+0.3,0.15);
 \node[anchor = west, font = \footnotesize\bfseries] at (\dx+0.4,0.15) {PR};
 \draw[green, thick] (2*\dx,0.15) -- (2*\dx+0.3,0.15);
 \node[anchor = west, font = \footnotesize\bfseries] at (2*\dx+0.4,0.15) {NPR};
\end{tikzpicture}

\caption{\textbf{Parameter sensitivity analysis on \textsc{EventVOT}.}
The left panel shows performance variations with respect to the temporal alignment weight~$\lambda_{1}$ (\textbf{TA Module}), 
while the right panel presents variations with respect to the spatial alignment weight~$\lambda_{2}$ (\textbf{SAOT Module}).}
\label{fig:ablation_param}
\end{figure}
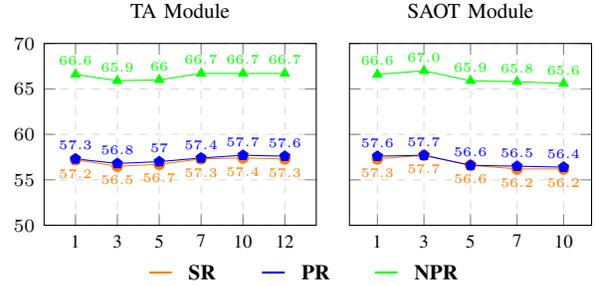

\begin{table}[!t]
	\centering
	\setlength{\tabcolsep}{8pt}
	\caption{\textbf{Ablation of cost distance definitions in the SAOT module.} \texttt{SAOT Distance} indicates the distance metric employed in constructing the optimal transport cost matrix.}
	\begin{tabular}{c|ccc|ccc}
	\toprule[1.1pt]
	\multirow{2}[2]{*}{\makecell{SAOT \\ Distance}} & 
	\multicolumn{3}{c|}{\textsc{EventVOT}} & 
	\multicolumn{3}{c}{\textsc{COESOT}} \\
	\cmidrule(lr){2-4} \cmidrule(lr){5-7}
	& SR & PR & NPR & SR & PR & NPR \\
	\midrule
	$\ell_{1\text{-}\cos}$ & 57.1 & 57.4 & 66.4 & 51.9 & 59.0 & 57.9 \\
	$\ell_1$ & 57.3 & 57.4 & 66.7 & 52.5 & 59.5 & 58.4 \\
	\rowcolor{gray!20}
	$\ell_2$ & \textbf{57.8} & \textbf{58.0} & \textbf{67.2} & \textbf{52.9} & \textbf{60.0} & \textbf{58.8} \\
	\bottomrule[1.1pt]
	\end{tabular}
	\label{tab:component_ablation_SAOT}
\end{table}
\begin{table}[!t]
	\centering
	\setlength{\tabcolsep}{5pt}
	\caption{\textbf{Ablation study of regularization strength $\varepsilon$.} Here, $\varepsilon$ represents the entropy regularization weight in the SAOT module, balancing transport cost and entropy to control stability, efficiency, and solution smoothness.}
	\begin{tabular}{c|c|ccc|ccc}
	\toprule[1.1pt]
	\multirow{2}[2]{*}{No.} & 
	\multirow{2}[2]{*}{\makecell{Regularization \\ Strength $\varepsilon$}} & 
	\multicolumn{3}{c|}{\textsc{EventVOT}} & 
	\multicolumn{3}{c}{\textsc{COESOT}} \\
	\cmidrule(lr){3-5} \cmidrule(lr){6-8}
	& & SR & PR & NPR & SR & PR & NPR \\
	\midrule
	1 & $1\mathrm{e}{+1}$ & 56.6 & 56.8 & 65.9 & 40.8 & 46.7 & 46.3 \\
	2 & $1\mathrm{e}{+0}$ & 56.4 & 56.6 & 65.8 & 51.4 & 57.7 & 56.8 \\
	3 & $1\mathrm{e}{-1}$ & 56.7 & 57.0 & 66.1 & 52.2 & 59.2 & 58.0 \\
	\rowcolor{gray!20}
	4 & $1\mathrm{e}{-2}$ & \textbf{57.8} & \textbf{58.0} & \textbf{67.2} & \textbf{52.9} & \textbf{60.0} & \textbf{58.8} \\
	\bottomrule[1.1pt]
	\end{tabular}
	\label{tab:reg_ablation}
\end{table}

\subsection{Ablation Studies}
\label{ablation_studies}
\begin{figure*}[!t]
	\centering
	\tabcolsep = 1pt
	\begin{tabular}{ccccc}
	\raisebox{3\height}{\footnotesize{(a)}} &
	\includegraphics[width = 0.22\linewidth, frame]{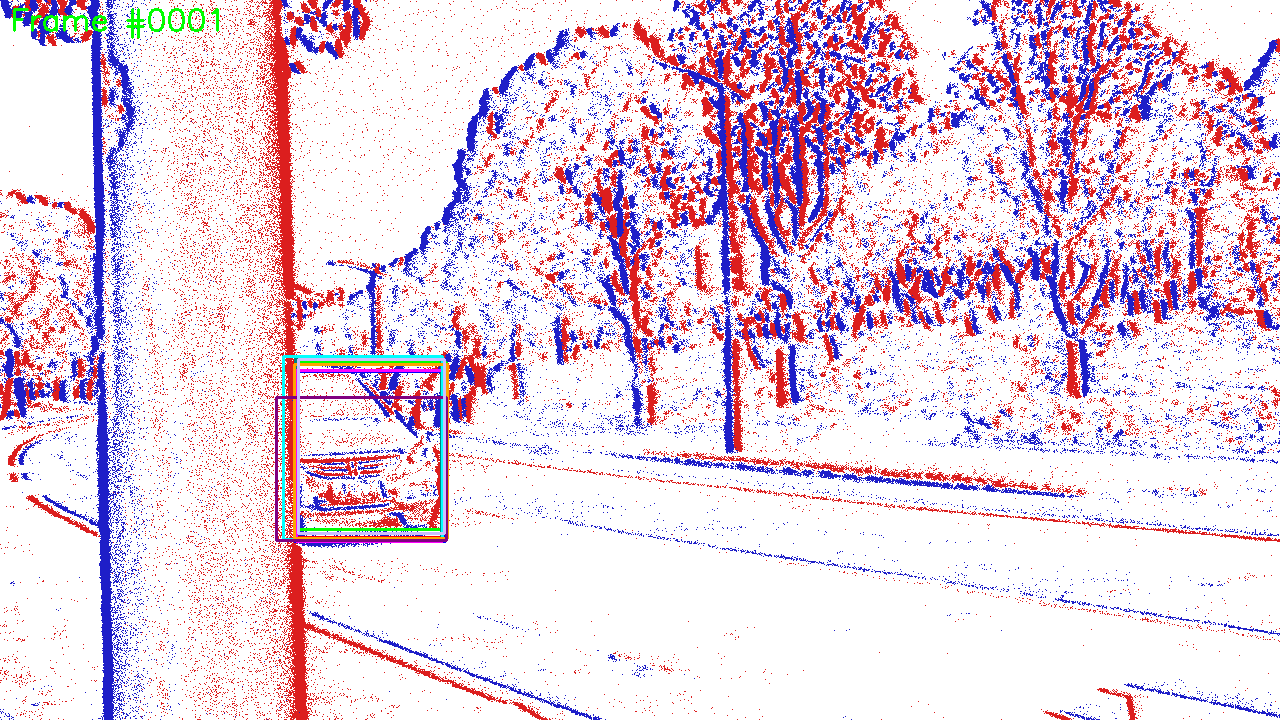} &
	\includegraphics[width = 0.22\linewidth, frame]{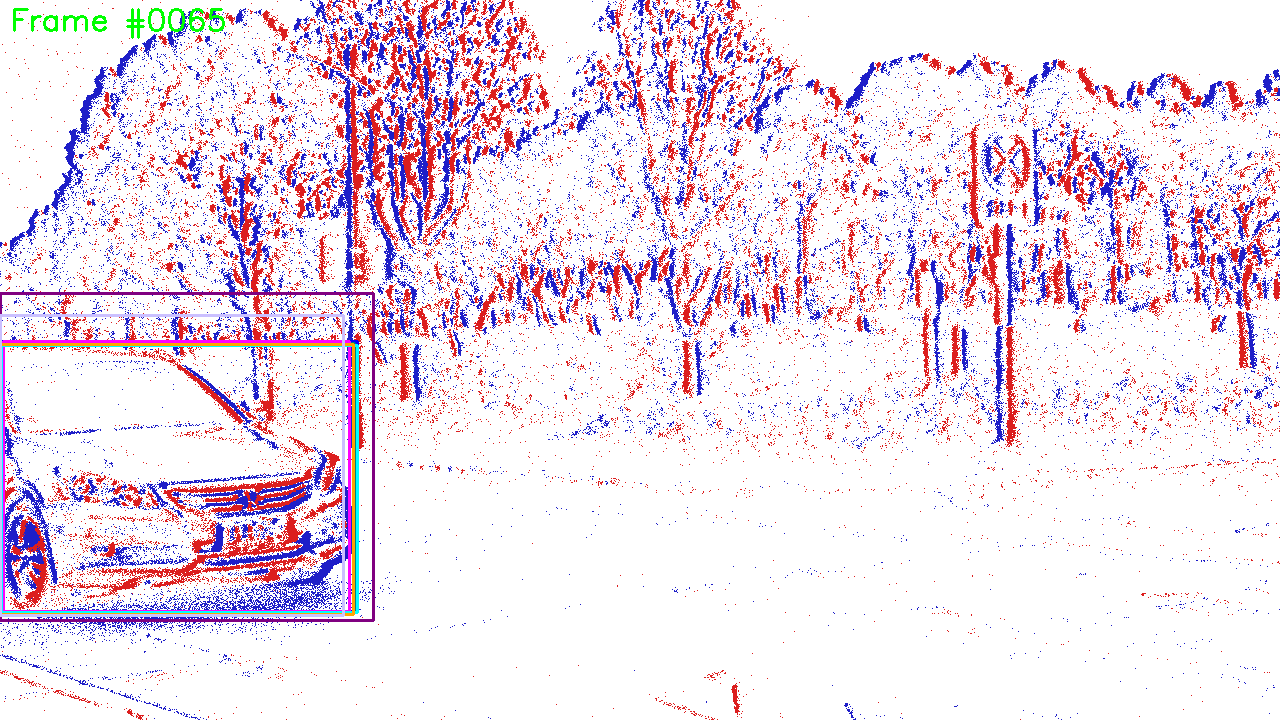} &
	\includegraphics[width = 0.22\linewidth, frame]{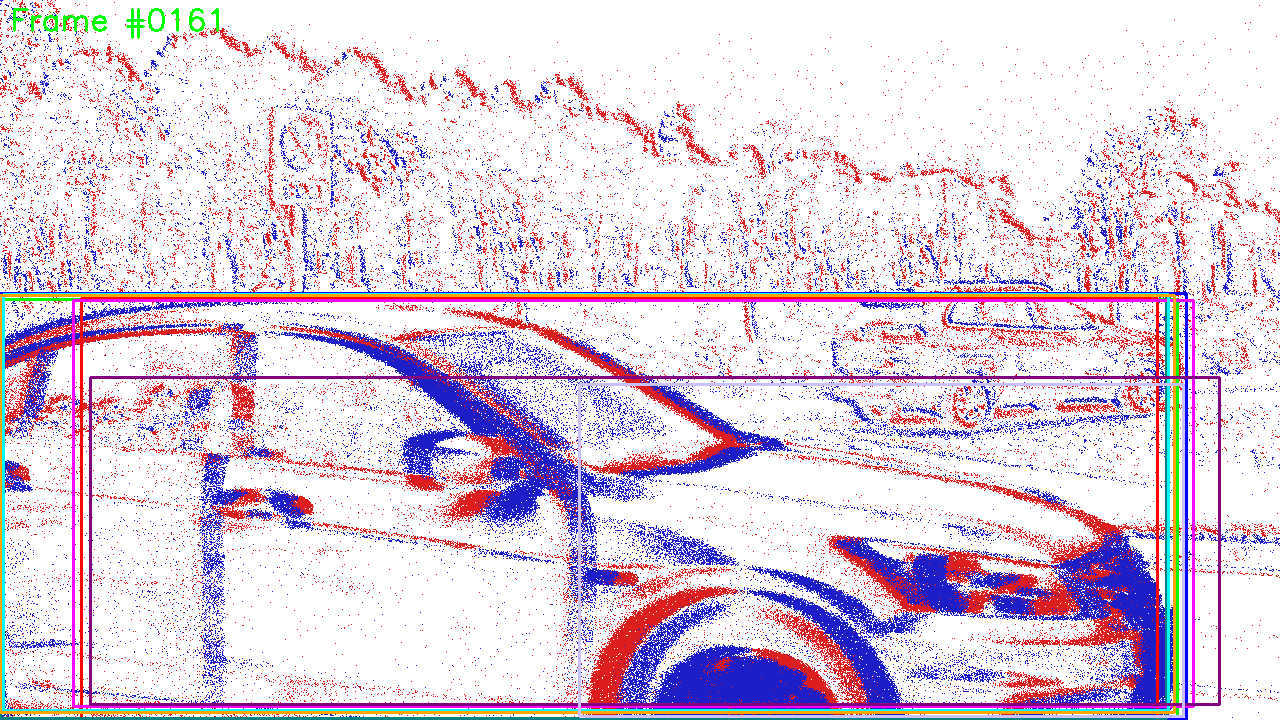} &
	\includegraphics[width = 0.22\linewidth, frame]{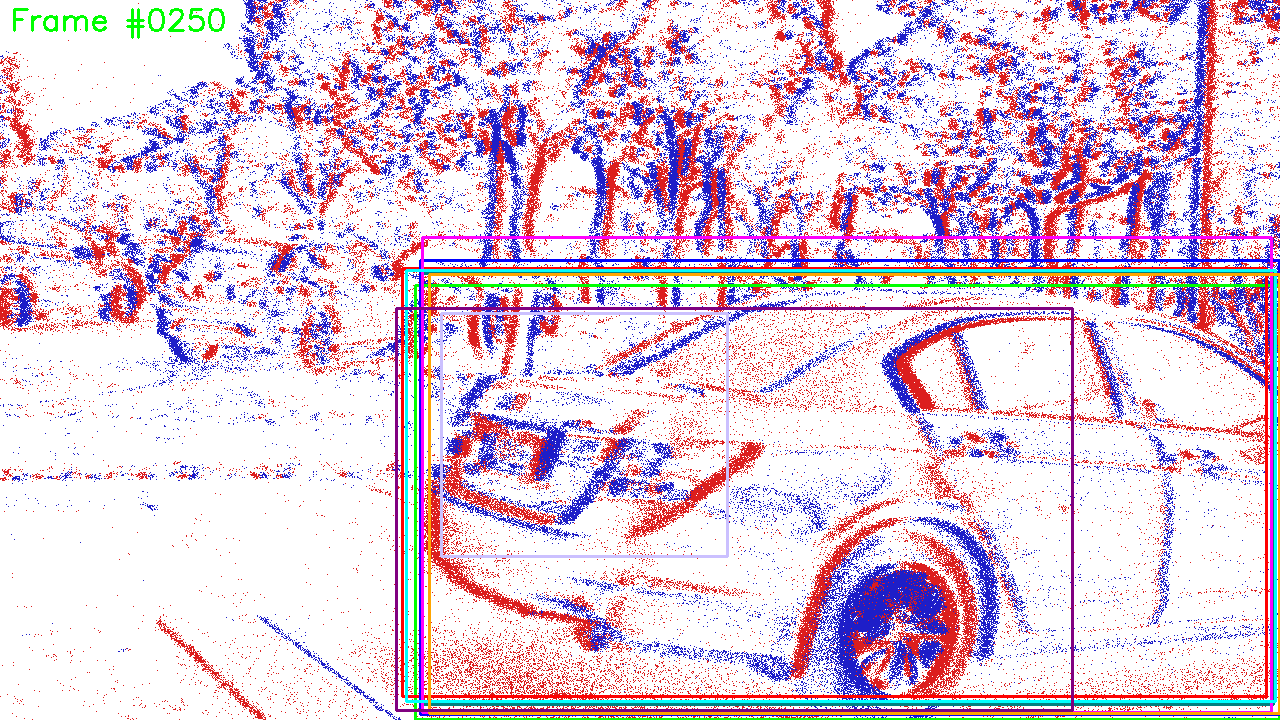} \\[-1pt]

	\raisebox{3\height}{\footnotesize{(b)}} &
	\includegraphics[width = 0.22\linewidth, frame]{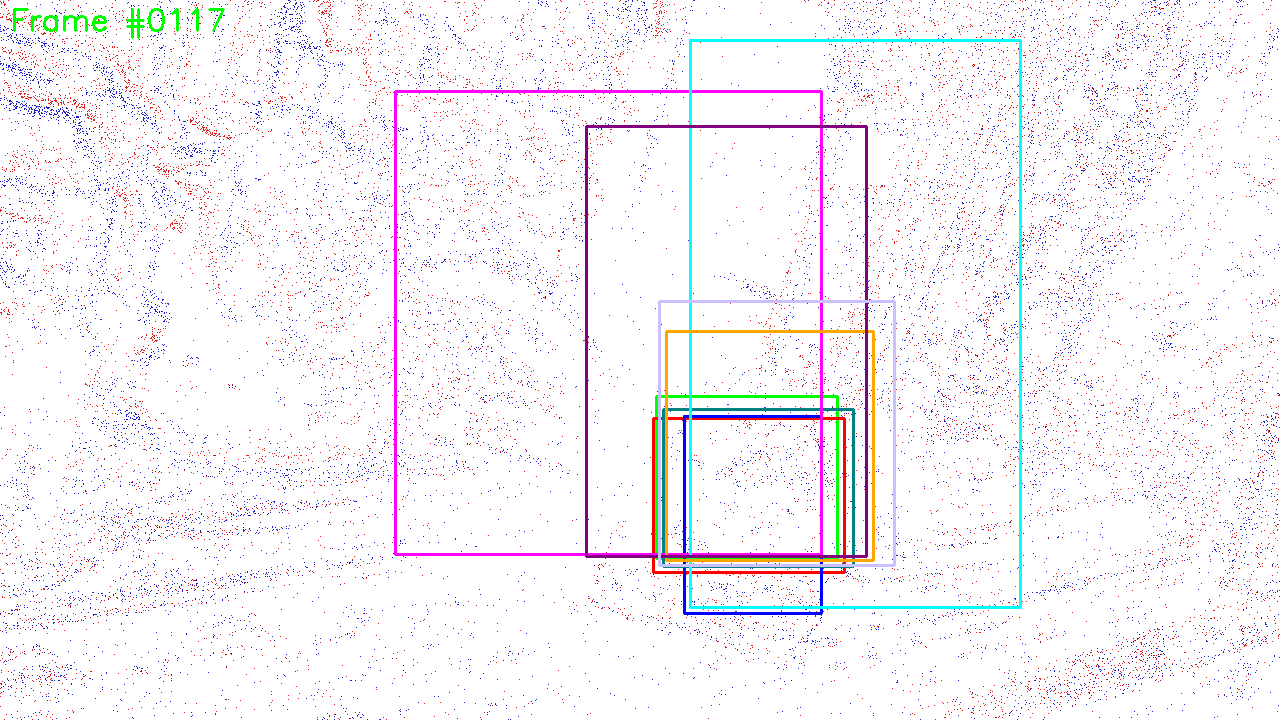} &
	\includegraphics[width = 0.22\linewidth, frame]{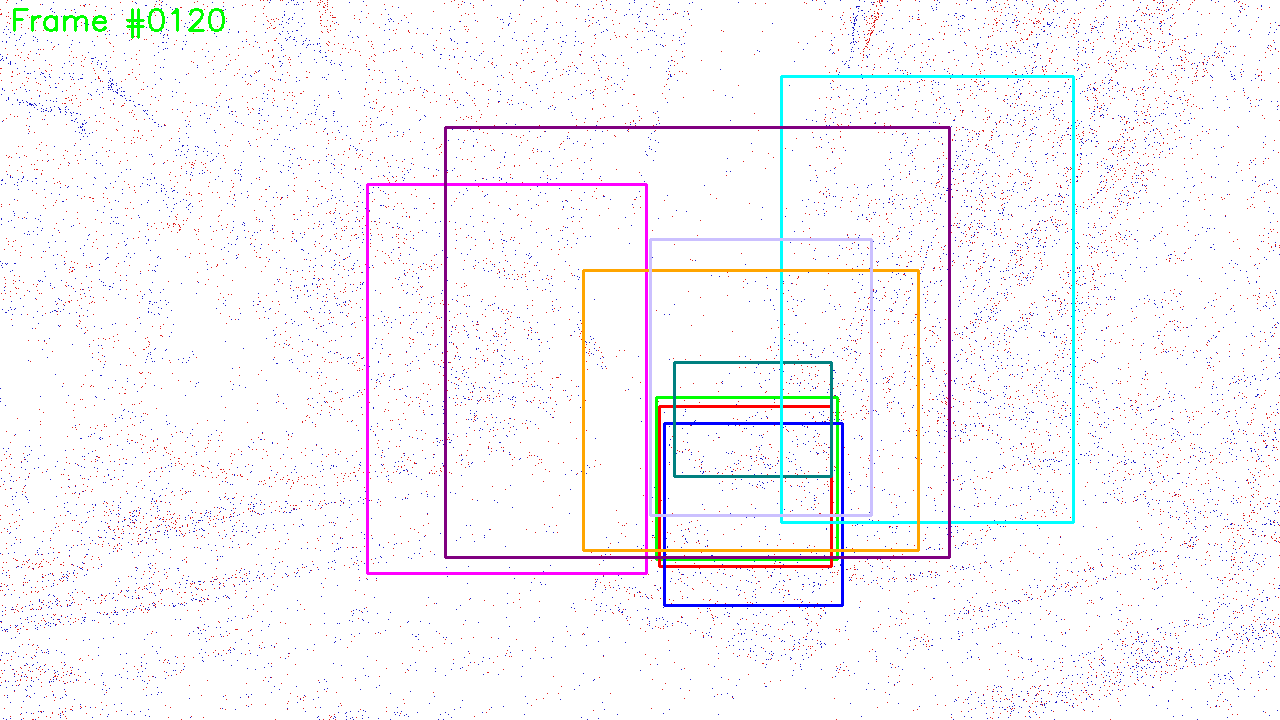} &
	\includegraphics[width = 0.22\linewidth, frame]{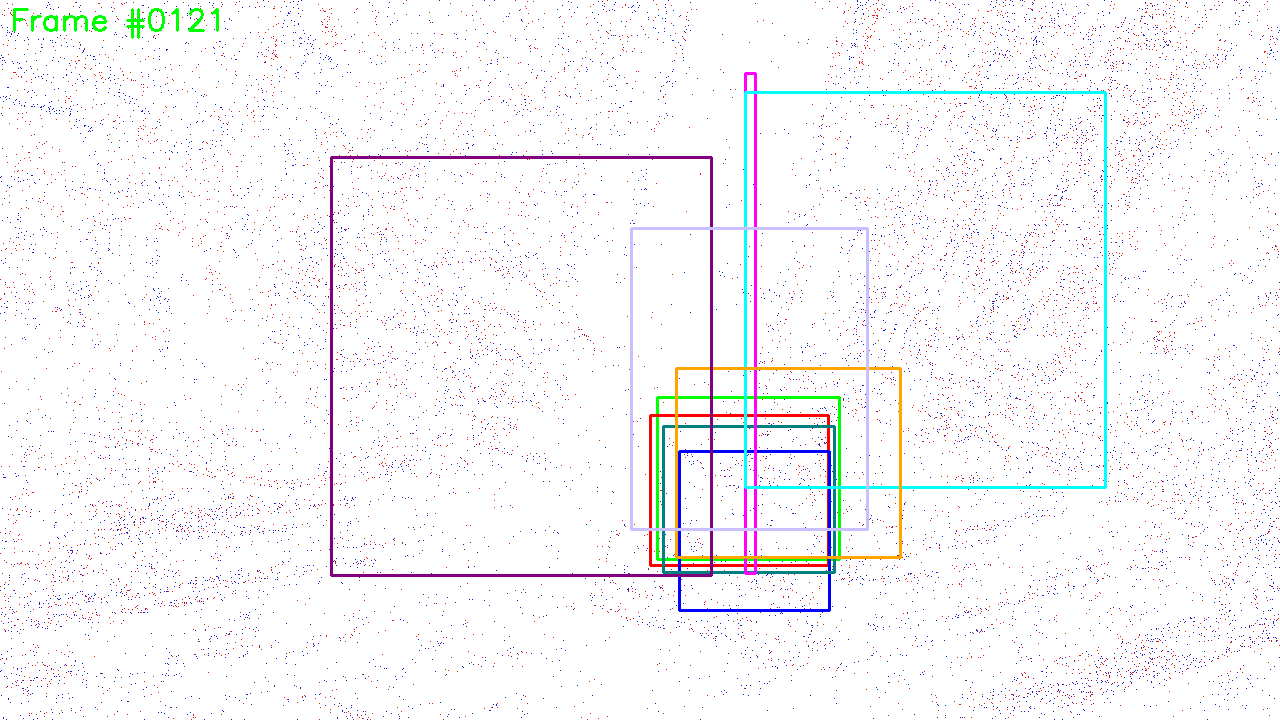} &
	\includegraphics[width = 0.22\linewidth, frame]{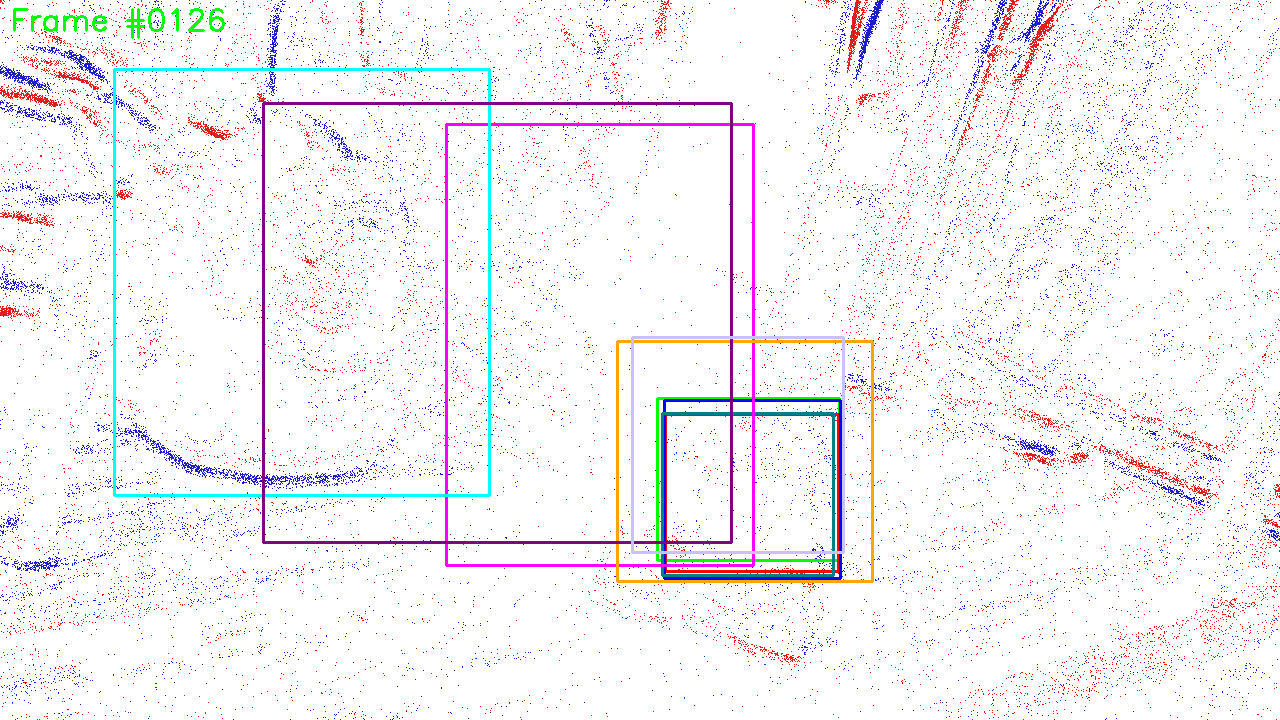} \\[-1pt]

	\raisebox{3\height}{\footnotesize{(c)}} &
	\includegraphics[width = 0.22\linewidth, frame]{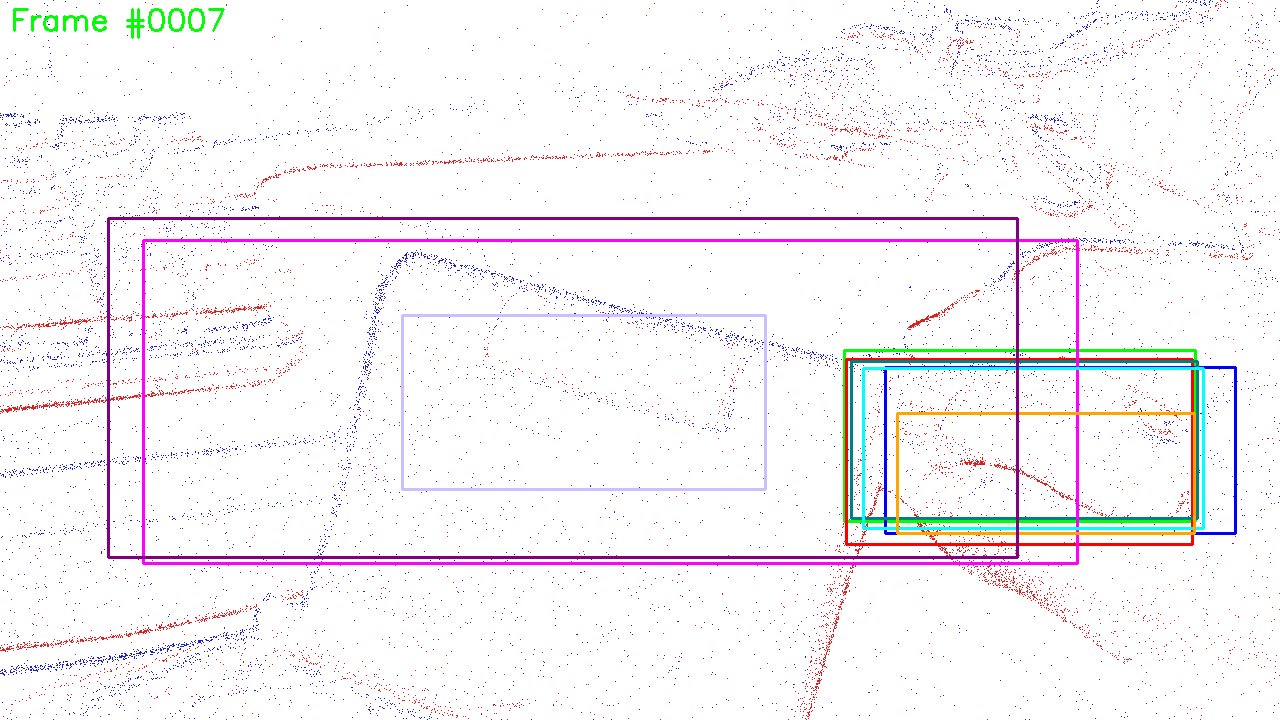} &
	\includegraphics[width = 0.22\linewidth, frame]{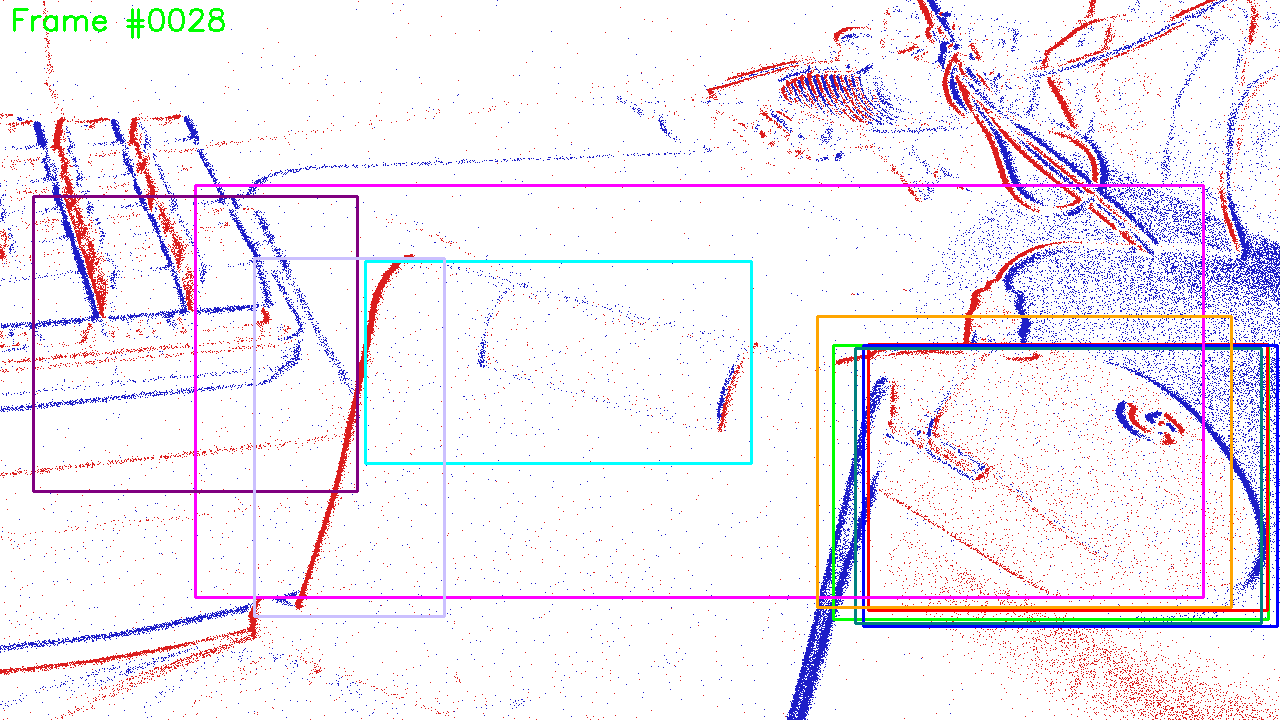} &
	\includegraphics[width = 0.22\linewidth, frame]{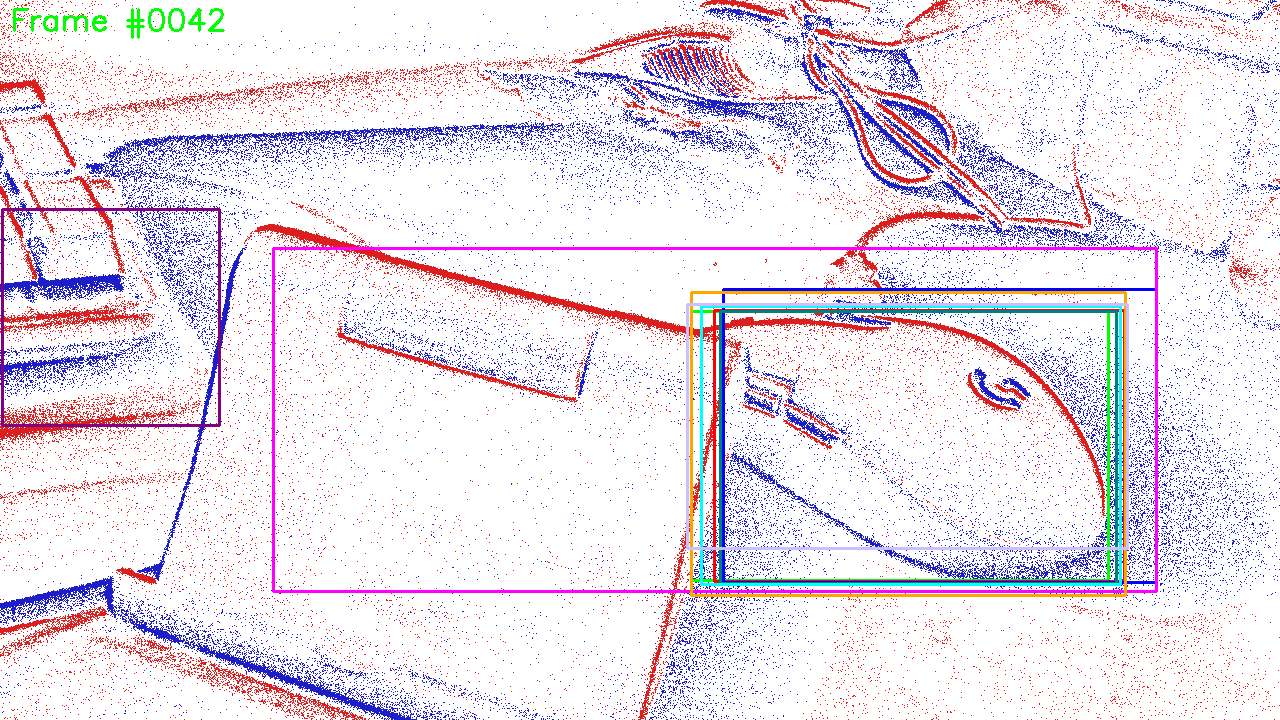} &
	\includegraphics[width = 0.22\linewidth, frame]{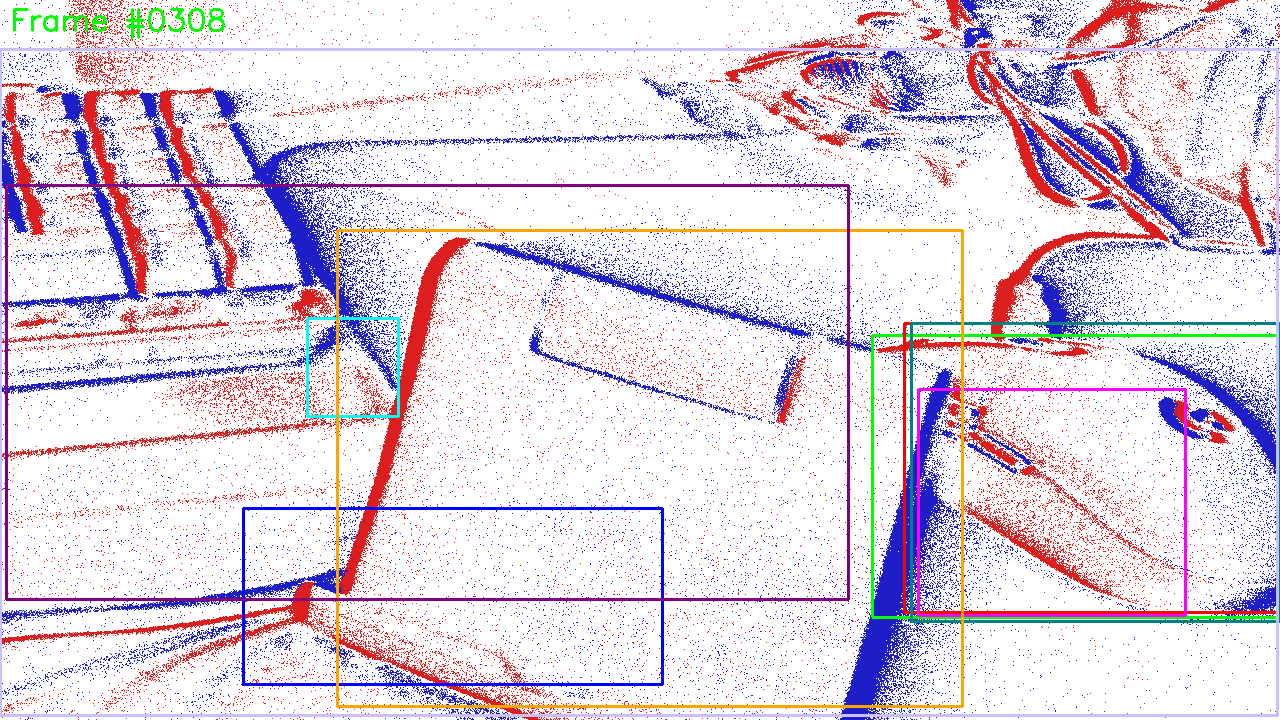} \\[-1pt]

	\raisebox{3\height}{\footnotesize{(d)}} &
	\includegraphics[width = 0.22\linewidth, frame]{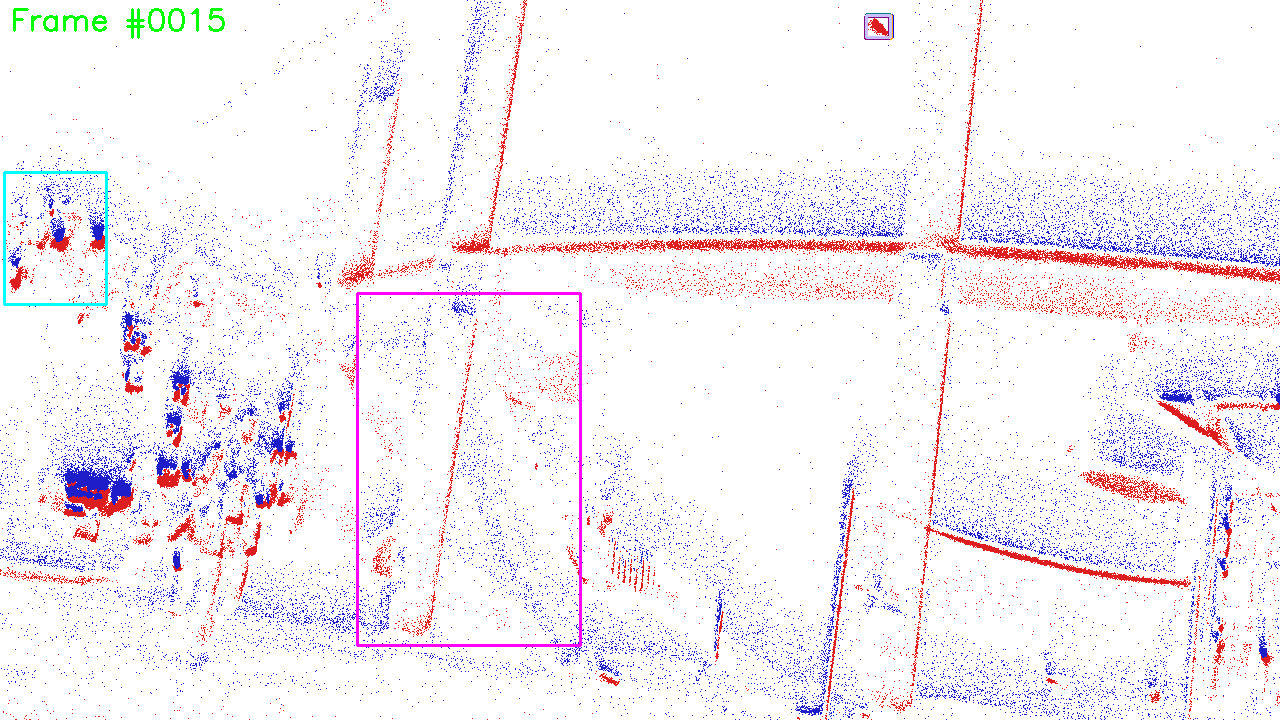} &
	\includegraphics[width = 0.22\linewidth, frame]{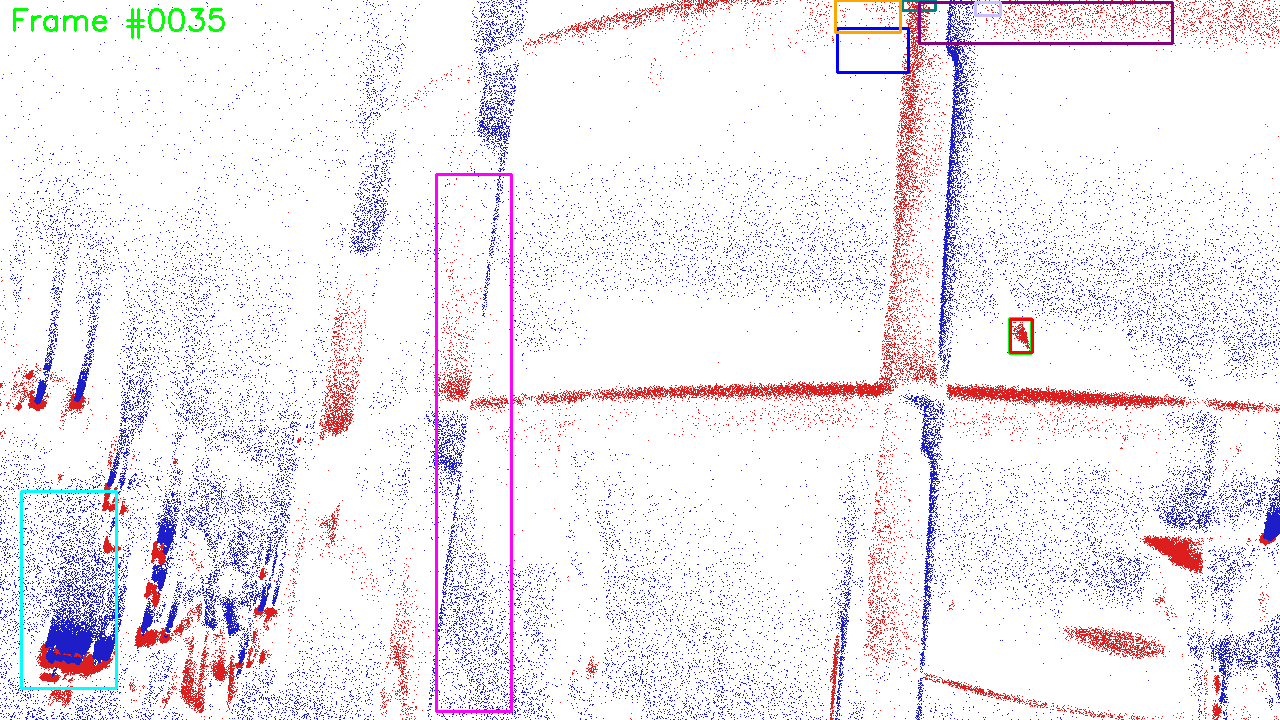} &
	\includegraphics[width = 0.22\linewidth, frame]{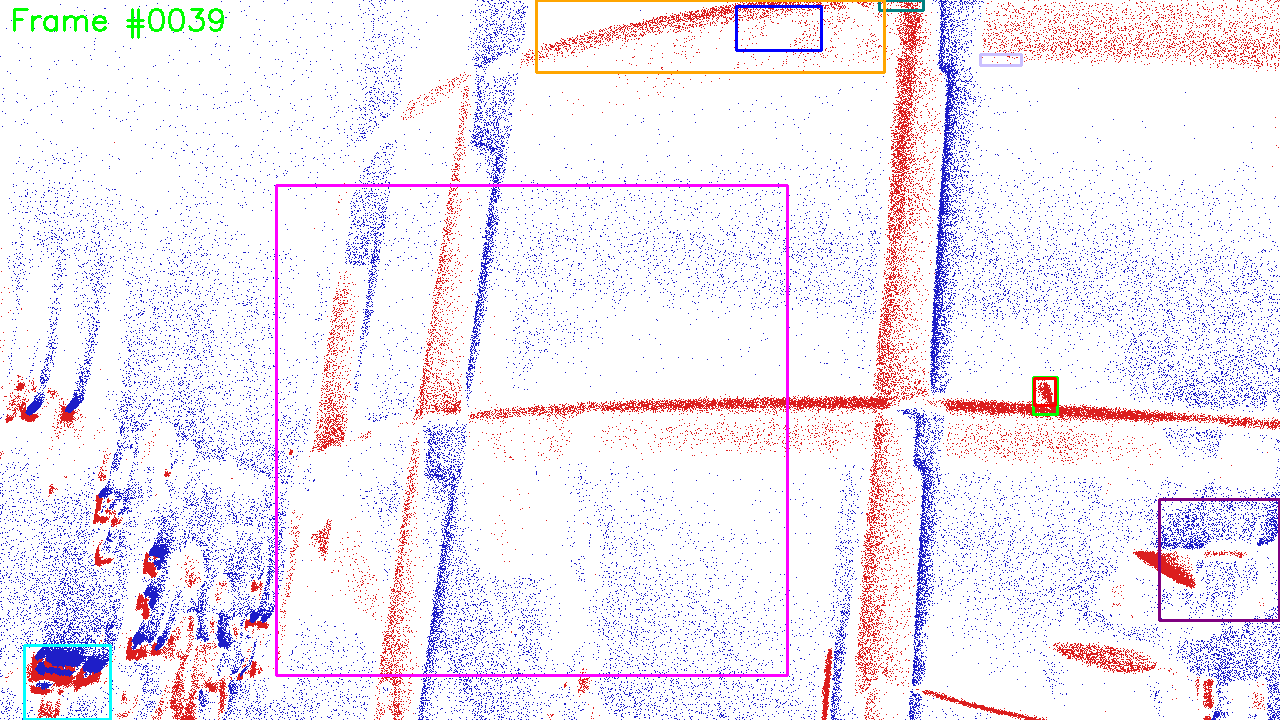} &
	\includegraphics[width = 0.22\linewidth, frame]{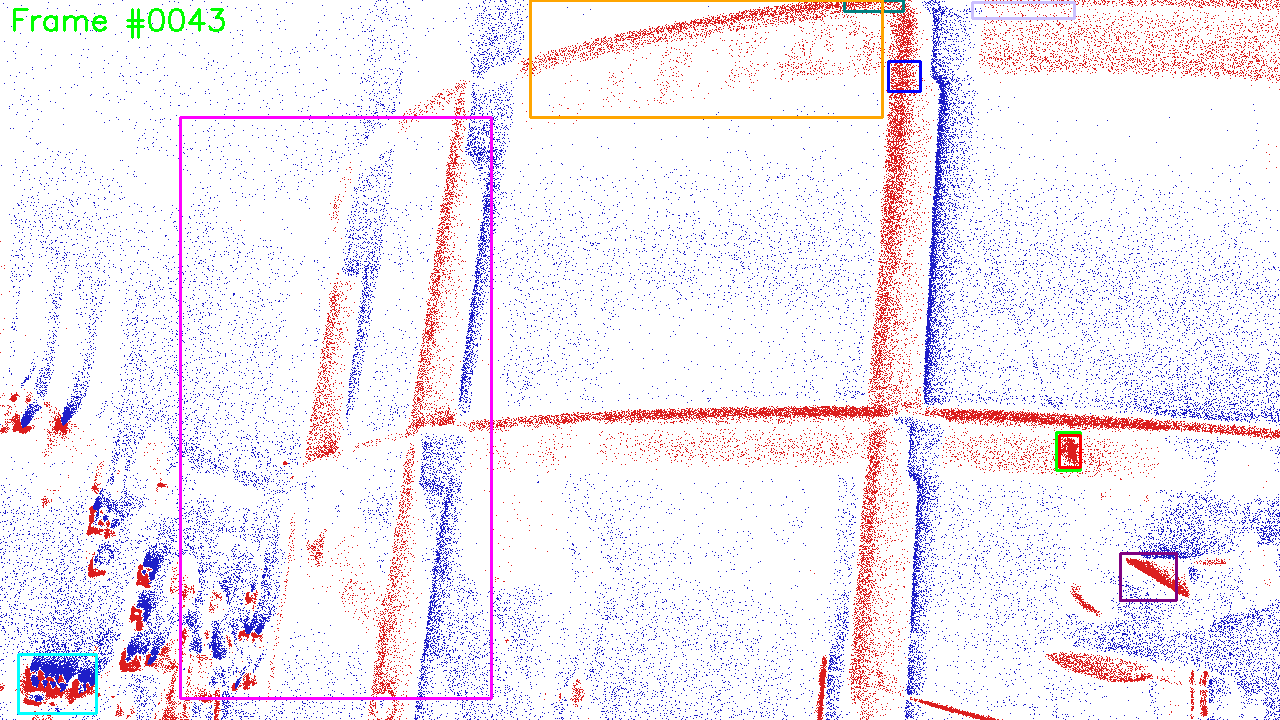} \\[-1pt]

	\raisebox{3\height}{\footnotesize{(e)}} &
	\includegraphics[width = 0.22\linewidth, frame]{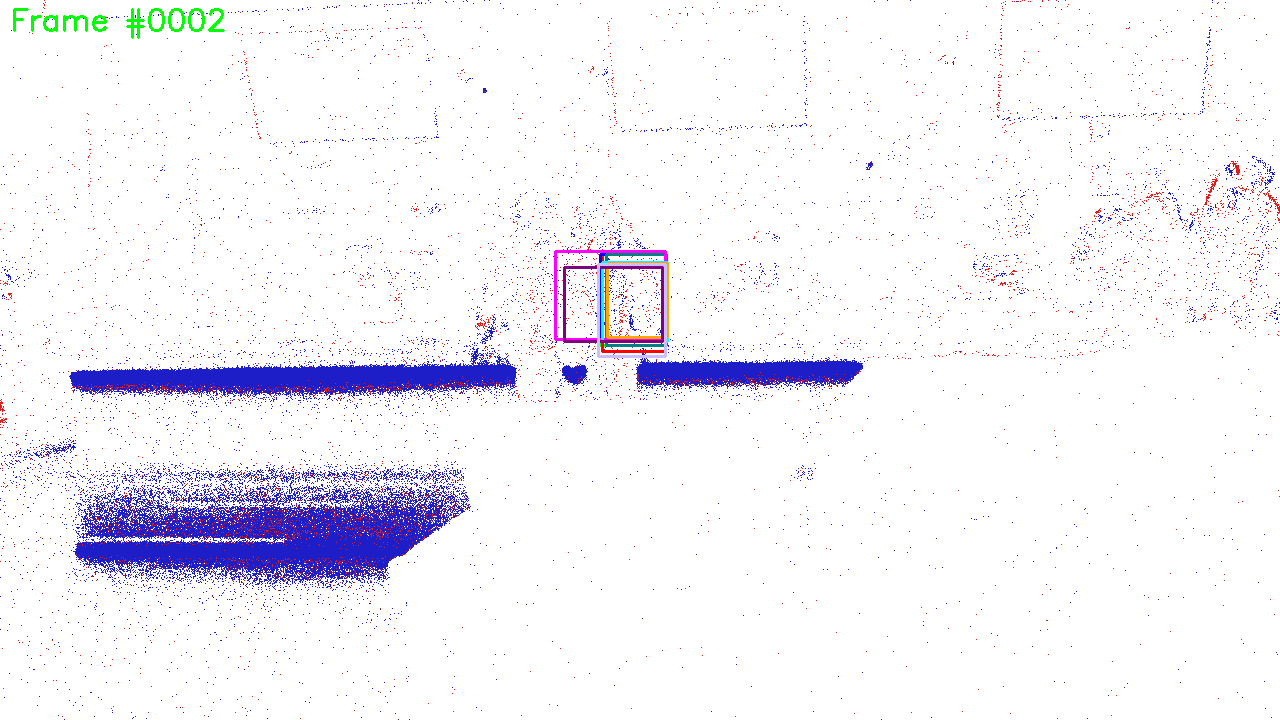} &
	\includegraphics[width = 0.22\linewidth, frame]{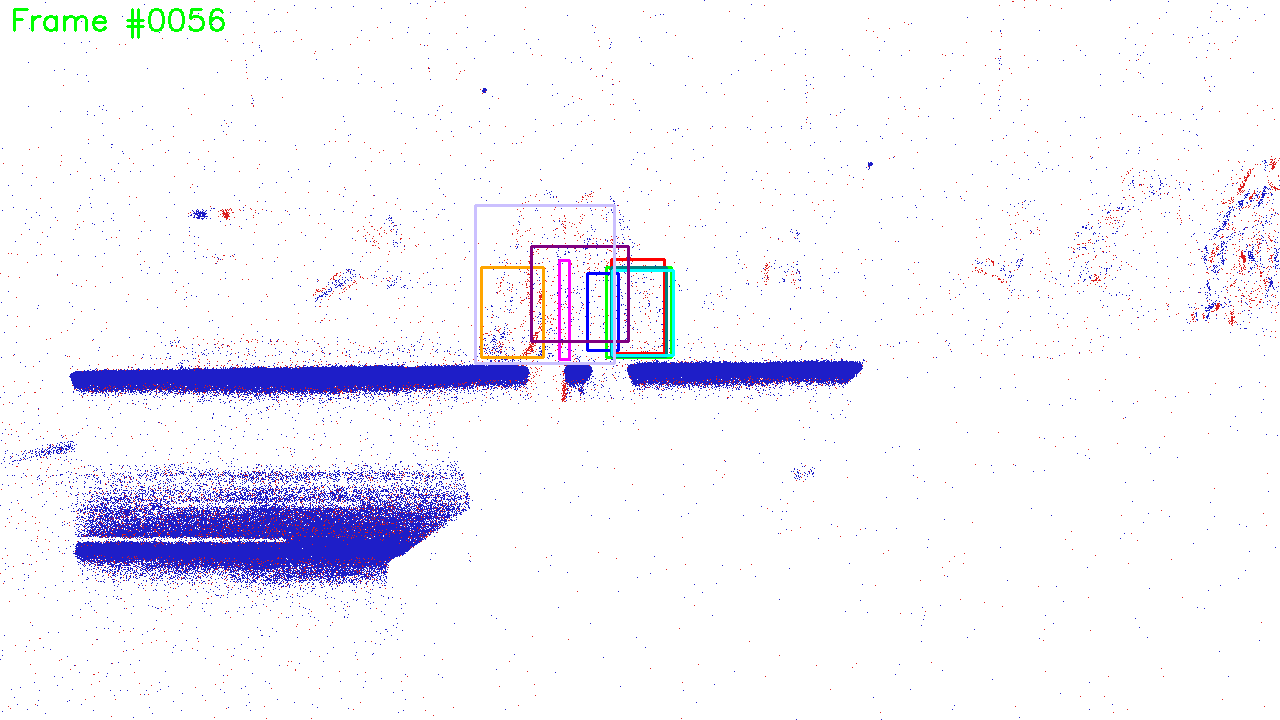} &
	\includegraphics[width = 0.22\linewidth, frame]{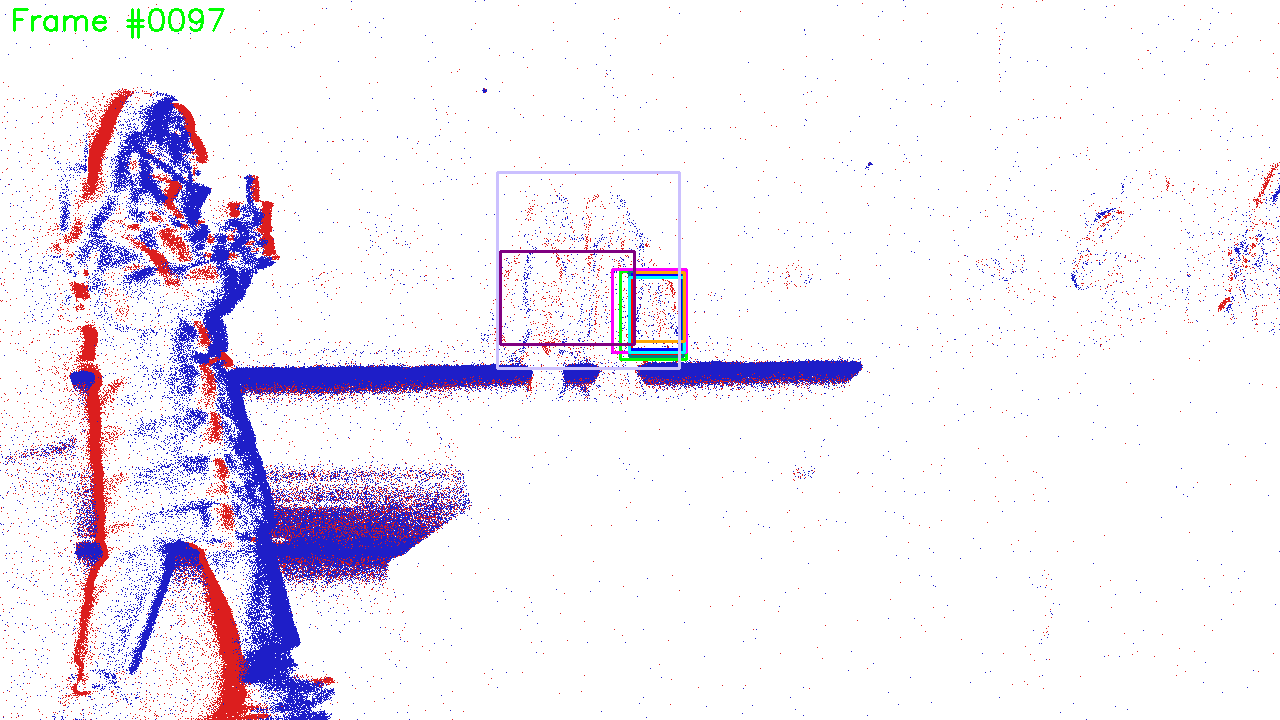} &
	\includegraphics[width = 0.22\linewidth, frame]{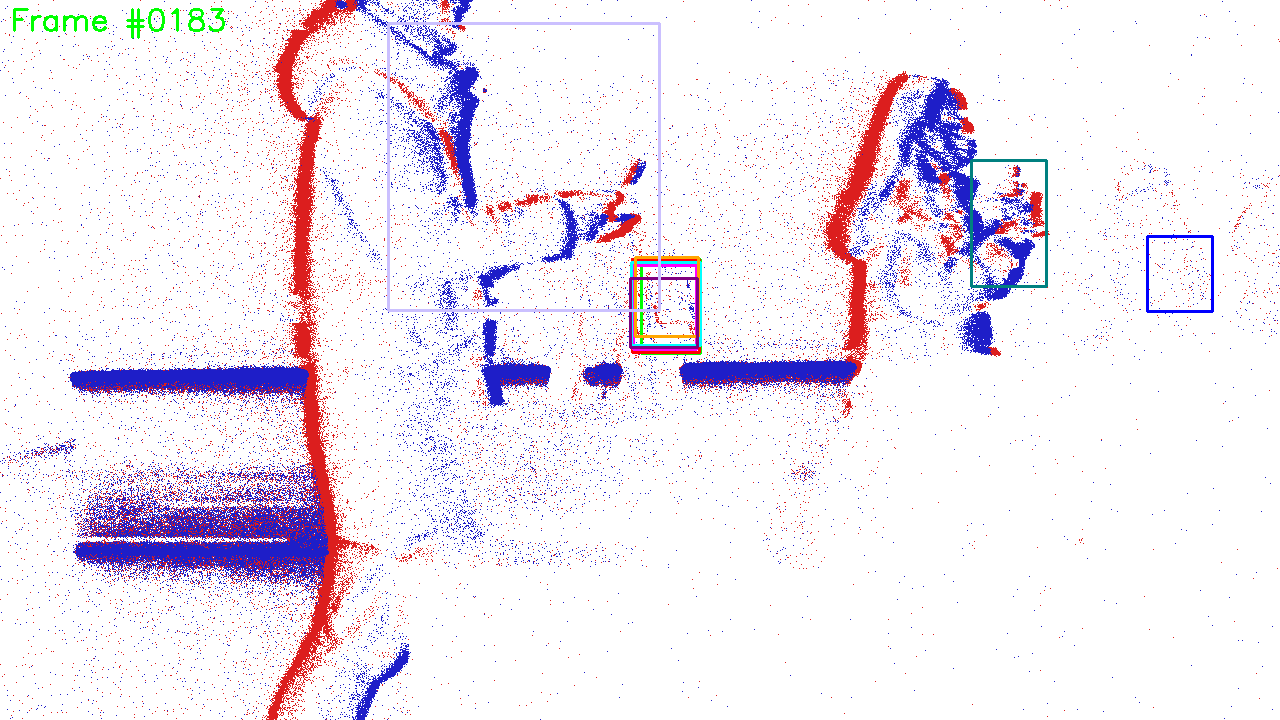} \\
	\end{tabular}

	% \vspace{3pt}
	\includegraphics[width = 0.8\linewidth]{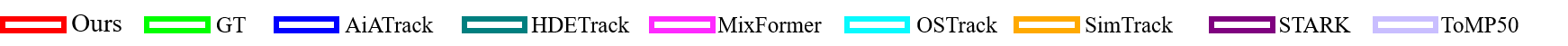}

	\caption{\textbf{Qualitative tracking results on \textsc{EventVOT}.} Examples illustrate HAD’s robustness under 
	(a) high-speed motion, 
	(b) sparse scenes, 
	(c) complex backgrounds, 
	(d) small objects, and 
	(e) occluded targets, 
	demonstrating stable and accurate target localization across diverse conditions.}
	\label{fig:event_tracking_visualization}
\end{figure*}

\subsubsection{Effectiveness of Components}

\cref{tab:component} presents detailed ablation results on \textsc{EventVOT}, \textsc{COESOT}, and \textsc{VisEvent}. Both the Temporal Alignment (TA) and Spatial-Aligned Optimal Transport (SAOT) modules consistently improve performance across all datasets, with their combination yielding the best overall results. The improvements are most significant on \textsc{EventVOT}, where TA increases SR by +0.9\% and SAOT adds +1.2\%, while gains on \textsc{COESOT} are more modest at +0.2\% and +0.1\%, respectively. On \textsc{VisEvent}, the enhancements remain steady yet smaller in magnitude: TA contributes +0.2\% in SR and SAOT adds +0.1\%, culminating in a total SR improvement of +0.6\% over the baseline. Similar patterns are observed for PR and NPR, demonstrating consistent robustness across evaluation metrics.

These variations highlight the differing characteristics of each dataset. \textsc{EventVOT}~\cite{wang2024event} emphasizes extreme conditions such as high-speed motion and rapid dynamics, where RGB–event misalignment is severe. TA alleviates temporal asynchrony through GRU-based modeling, while SAOT compensates for motion-induced spatial distortion, hence the larger gains. In contrast, \textsc{COESOT}~\cite{tang2022coesot} and \textsc{VisEvent}~\cite{wang2021viseventbenchmark} focus on occlusion, clutter, and scale variation, where the baseline already performs robustly; thus, explicit alignment provides only marginal additional benefit.

Overall, these findings confirm that HAD’s modules not only effectively address the core spatio-temporal asymmetry motivating our design but also generalize well across diverse tracking benchmarks, from high-speed event streams to low-light visual sequences, demonstrating strong versatility and robustness.

\subsubsection{Effectiveness of TA Variants}

To identify the optimal TA design, we compared RNN, GRU, Bi-GRU, Bi-LSTM, Mamba, and an MLP baseline (see \cref{tab:component_ablation_TA}). GRU achieves the best trade-off: on \textsc{EventVOT}, SR~57.8\%, PR~58.0\%, NPR~67.2\%; on \textsc{COESOT}, SR~52.9\%, PR~60.0\%, NPR~58.8\%. Bidirectional models (\textit{e.g.}, Bi-LSTM) approach GRU in accuracy but incur higher cost, while shallow models (RNN, Mamba, MLP) underperform. For instance, MLP attains only SR~51.8\% on \textsc{COESOT}. 

GRU excels because its gating mechanism captures long-range dependencies in sparse, asynchronous streams without gradient vanishing, while its unidirectional structure preserves causal consistency and avoids noise overfitting. Compared with global-context models (\textit{e.g.}, Mamba), GRU’s recurrent design better maintains local temporal coherence, enabling precise alignment of fast-moving targets.

\subsubsection{Effectiveness of Cost Metrics}

\cref{tab:component_ablation_SAOT} compares cost metrics in SAOT. $\ell_2$ consistently outperforms $\ell_{1\text{-}\cos}$ and $\ell_1$ across SR, PR, and NPR. Its quadratic term amplifies subtle feature differences and stabilizes gradients, providing smoother convergence. In contrast, $\ell_{1\text{-}\cos}$ ignores absolute positional shifts, and $\ell_1$ lacks local sensitivity. Hence, $\ell_2$ best balances global adaptability with local precision, supporting accurate alignment under occlusion and scale variation.

\subsubsection{Effectiveness of Regularization Strength}

\cref{tab:reg_ablation} evaluates the impact of the Sinkhorn entropy regularization coefficient $\varepsilon$. Smaller $\varepsilon$ values yield the best results, while larger ones degrade performance. Weak regularization preserves fine-grained structure, allowing the solution to approach the unregularized optimum and remain sensitive to local variations. This validates our choice of small $\varepsilon$ for robust alignment, addressing motion blur and fine-scale distortions.

\subsubsection{Sensitivity to Alignment Weights}
\label{sec:sensitivity_to_alignment_weights}

\cref{fig:ablation_param} shows performance trends for different alignment weights $\lambda_1$ (TA) and $\lambda_2$ (SAOT). Optimal results occur at $\lambda_1 = 10$ (SR~57.4\%, PR~57.7\%, NPR~66.7\%) and $\lambda_2 = 3$ (SR~57.7\%, PR~57.7\%, NPR~67.0\%), suggesting that stronger temporal and moderate spatial alignment complement each other. This confirms the necessity of balanced supervision between temporal and spatial cues, two pillars of our asymmetry-motivated design.

\subsection{Qualitative Analysis}

\cref{fig:event_tracking_visualization} visualizes tracking on \textsc{EventVOT}. HAD maintains stable trajectories under high-speed motion, cluttered backgrounds, and small targets, highlighting TA’s role in temporal stabilization and SAOT’s role in spatial refinement. Even without RGB at inference, the student inherits rich spatial cues distilled during training, effectively addressing modality asymmetry.

\begin{figure*}[!t]
	\centering
	\tabcolsep = 1pt
	\begin{tabular}{ccccccc}
	\rotatebox{90}{\parbox{2.5cm}{\centering \footnotesize \textbf{HDETrack}~\cite{wang2024event}}} &
	\includegraphics[width = 0.15\linewidth, frame]{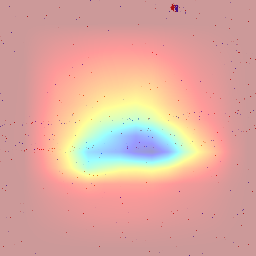} &
	\includegraphics[width = 0.15\linewidth, frame]{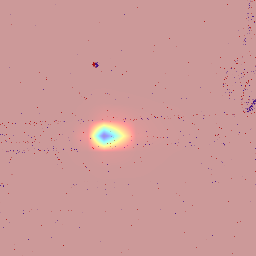} &
	\includegraphics[width = 0.15\linewidth, frame]{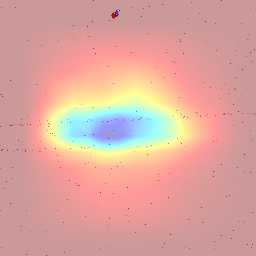} &
	\includegraphics[width = 0.15\linewidth, frame]{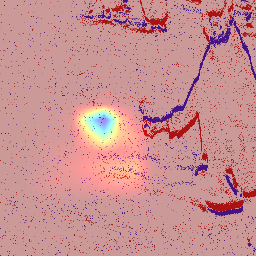} &
	\includegraphics[width = 0.15\linewidth, frame]{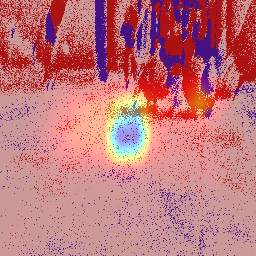} &
	\includegraphics[width = 0.15\linewidth, frame]{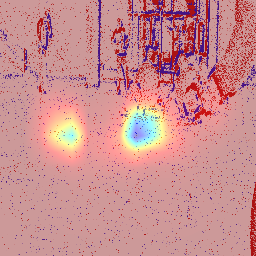} \\[-1pt]

	\rotatebox{90}{\parbox{2.5cm}{\centering \footnotesize \textbf{Ours}}} &
	\includegraphics[width = 0.15\linewidth, frame]{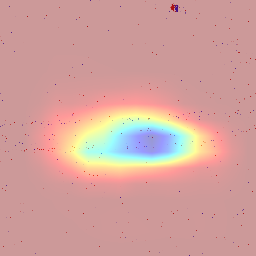} &
	\includegraphics[width = 0.15\linewidth, frame]{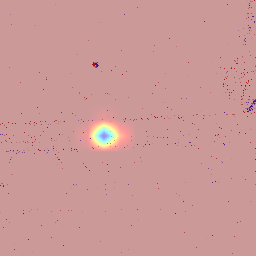} &
	\includegraphics[width = 0.15\linewidth, frame]{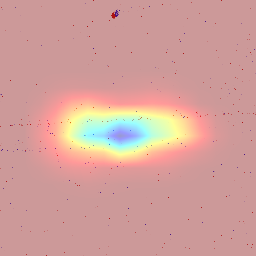} &
	\includegraphics[width = 0.15\linewidth, frame]{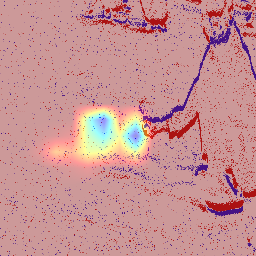} &
	\includegraphics[width = 0.15\linewidth, frame]{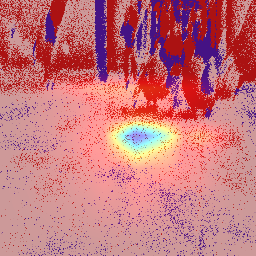} &
	\includegraphics[width = 0.15\linewidth, frame]{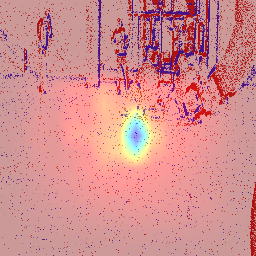} \\[-1pt]

	\rotatebox{90}{\parbox{2.5cm}{\centering \footnotesize \textbf{Teacher}}} &
	\includegraphics[width = 0.15\linewidth, frame]{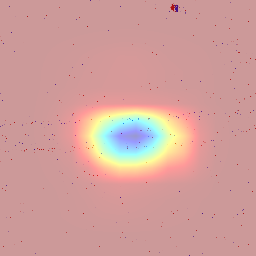} &
	\includegraphics[width = 0.15\linewidth, frame]{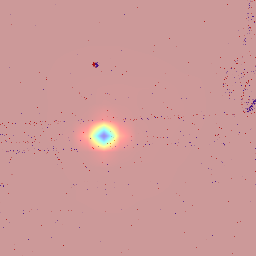} &
	\includegraphics[width = 0.15\linewidth, frame]{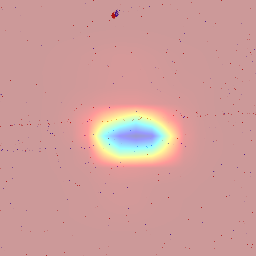} &
	\includegraphics[width = 0.15\linewidth, frame]{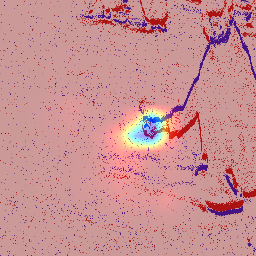} &
	\includegraphics[width = 0.15\linewidth, frame]{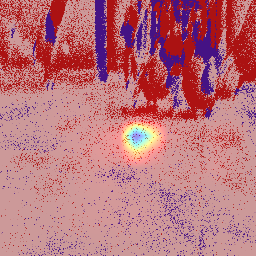} &
	\includegraphics[width = 0.15\linewidth, frame]{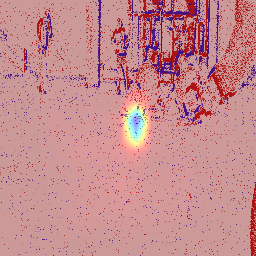} \\
	\end{tabular}
	\caption{\textbf{Visualization of response map comparisons on \textsc{COESOT}.} \textbf{Teacher} denotes the bimodal teacher network, while \textbf{Ours} and \textbf{HDETrack} represent event-only student networks trained with their respective distillation strategies. Our method produces response maps more closely aligned with the teacher, indicating improved spatial consistency and cross-modal knowledge transfer.}
	\label{fig:response_map}
\end{figure*}

\cref{fig:response_map} shows that student response maps distilled by HAD closely match the dual-modal teacher in hotspot distribution and intensity, demonstrating effective knowledge transfer and accurate target localization. Compared with HDETrack, HAD preserves sharper responses in cluttered scenes, underscoring enhanced robustness.

\begin{figure}[!t]
	\centering
	\tabcolsep = 3pt
	\begin{tabular}{ccc} 
	\includegraphics[width = 0.3\linewidth, frame]{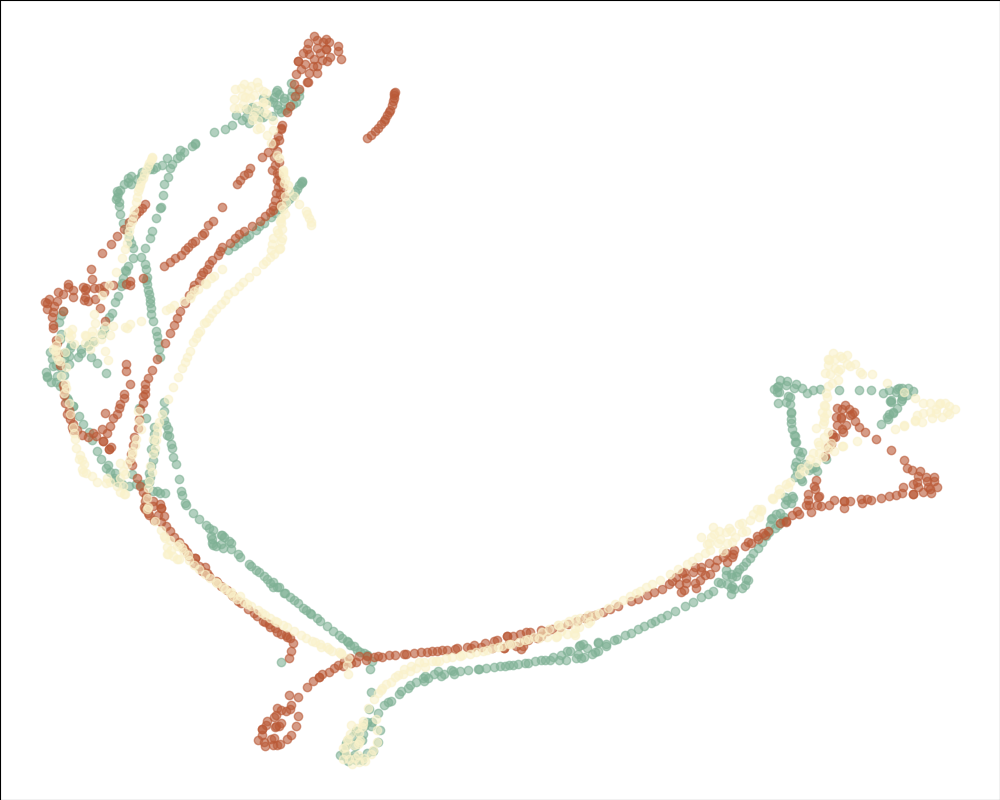} & 
	\includegraphics[width = 0.3\linewidth, frame]{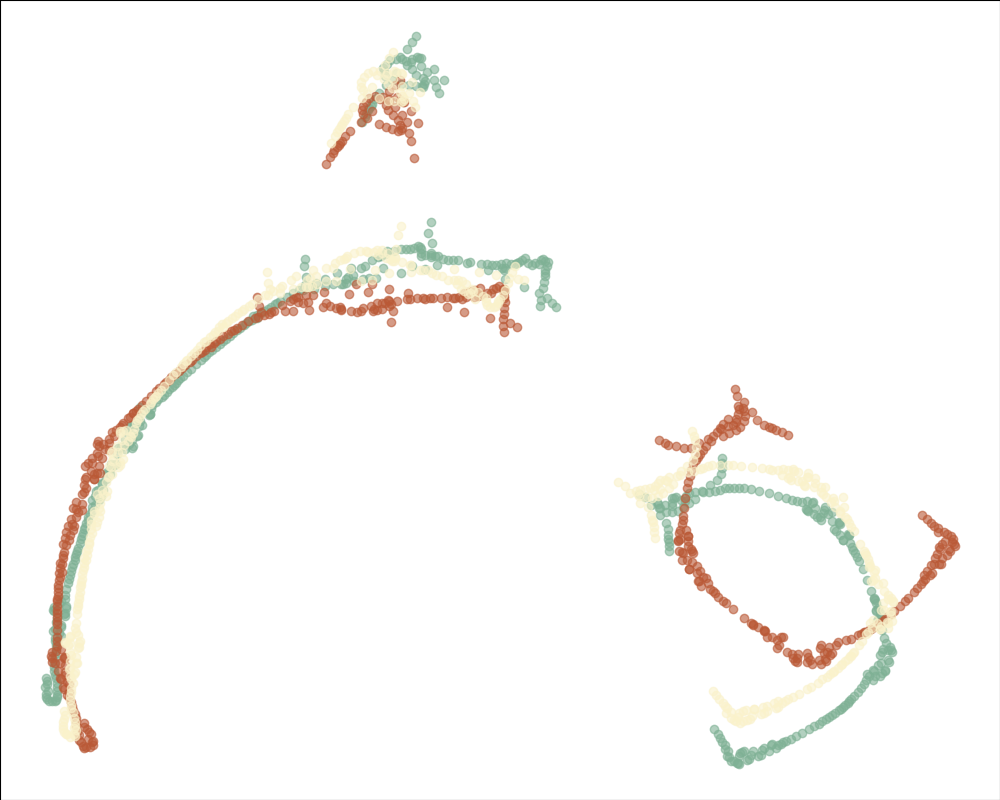} & 
	\includegraphics[width = 0.3\linewidth, frame]{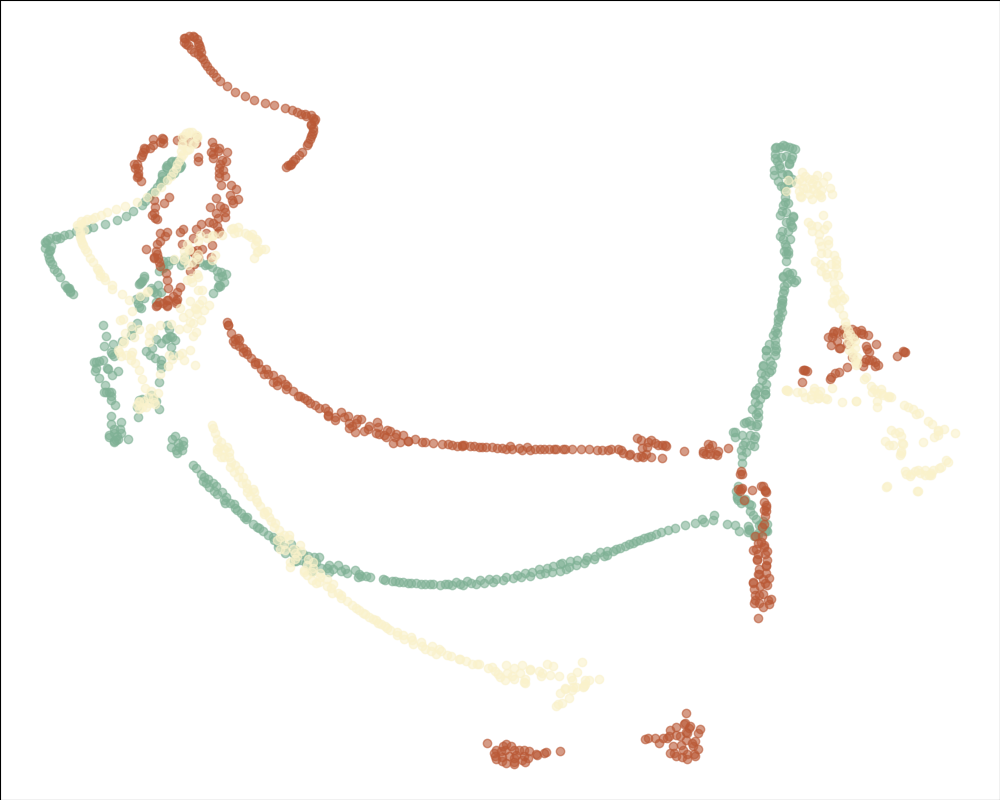}
	\end{tabular}

	% \vspace{3pt}
	\includegraphics[width = 0.4\linewidth]{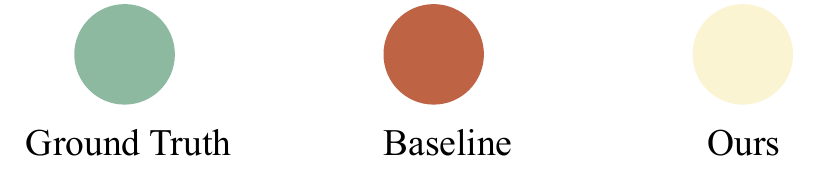}

	\caption{\textbf{t-SNE visualization of predicted bounding-box embeddings on \textsc{EventVOT}.} HAD produces tighter and more coherent clusters that align closely with the ground truth, demonstrating superior spatial consistency and feature separability compared with the baseline.}
	\label{fig:tsne}
\end{figure}

Finally, \cref{fig:tsne} presents t-SNE embeddings of predicted bounding boxes. HAD yields compact, well-separated clusters, confirming improved feature alignment and spatial consistency. These qualitative results further substantiate HAD’s effectiveness in mitigating spatio-temporal asymmetry.

\section{Conclusion and Limitation}

In this work, we identified and formalized the spatio-temporal asymmetry between RGB frames and event streams in single-object tracking (SOT). To address this challenge, we proposed {Hierarchical Asymmetric Distillation} (HAD), which integrates a GRU-based temporal alignment module and an entropic optimal transport-based spatial alignment module within the distillation framework. By explicitly bridging modality gaps, HAD enables a unimodal student network to inherit knowledge from a bimodal teacher without increasing model complexity. Extensive experiments on \textsc{EventVOT} and \textsc{COESOT} demonstrate that HAD significantly enhances robustness and accuracy under low-light, high-speed, and cluttered conditions, achieving state-of-the-art performance and validating its effectiveness against the core challenges motivating our design.

Despite these advantages, HAD has two main limitations. First, although it effectively mitigates spatio-temporal misalignment between RGB and event modalities and improves data efficiency, its applicability to other modalities, such as optical flow or depth, remains unexplored. Second, tracking accuracy in complex scenarios can still be improved. Future work will focus on: \textit{1)} extending HAD to additional modalities through unified fusion architectures and cross-modal learning paradigms, and \textit{2)} enhancing robustness via multi-scale feature refinement.

\bibliographystyle{IEEEtran}
\bibliography{HAD}

\vfill

\end{document}